%% file: main.tex
\documentclass{article} % For LaTeX2e
\usepackage{iclr2025_conference,times}

\input{math_commands.tex}

\usepackage{hyperref}
\usepackage{url}

\iclrfinalcopy % Uncomment for camera-ready version, but NOT for submission.
\usepackage{siunitx}
\usepackage{graphicx}%
\usepackage{multirow}%
\usepackage{amsmath,amssymb,amsfonts}%
\usepackage{amsthm}%
\usepackage{mathrsfs}%
\usepackage[title]{appendix}%
\usepackage{textcomp}%
\usepackage{manyfoot}%
\usepackage{booktabs}%
\usepackage{listings}%
\usepackage{xspace}
\usepackage{tabularx}
\usepackage{enumitem}

\usepackage[ruled,vlined]{algorithm2e}

\usepackage[utf8]{inputenc} % allow utf-8 input
\usepackage[T1]{fontenc}    % use 8-bit T1 fonts
\usepackage{hyperref}       % hyperlinks
\usepackage{url}            % simple URL typesetting
\usepackage{booktabs}       % professional-quality tables
\usepackage{amsfonts}       % blackboard math symbols
\usepackage{nicefrac}       % compact symbols for 1/2, etc.
\usepackage{microtype}      % microtypography
\usepackage{xcolor}         % colors

\usepackage{subfiles}
\usepackage{chapterbib}
\input{00-00-definitions}

\title{A Computational Framework for Modeling \\Emergence of Color Vision in the Human Brain}

\author{Atsunobu Kotani \& Ren Ng\\
Department of Electrical Engineering and Computer Sciences\\
University of California, Berkeley\\
\texttt{\{akotani,ren\}@berkeley.edu}
}

\begin{document}

% \newpage
% \input{00-00-todo.tex}

\maketitle

\input{00-Abstract}

\input{01-Introduction}

\input{02-Related_Work}
\input{03-Simulation_Engine}
\input{04-Cortex}

\input{05-Representation}

\input{06-Results}

\input{07-Conclusion}
\input{08-Acknowledgement}

\bibliography{bibliography}

\newpage
\nocite{*}

\part*{Supplementary Materials}
% % \input{S-00-supplement-main}
\setcounter{section}{0}
\renewcommand\thesection{S\arabic{section}}

\appendix
\input{S-01-intro}
\input{S-02-retina-model}
\input{S-03-cortical-model}
\input{S-04-CMF-SIM}

\input{S-05-baseline}

\input{S-06-gene-therapy}

\input{S-07-tetrachromacy}

\input{Rebuttal/RS-08-Extensive_Results}
\input{Rebuttal/RS-09-Table_of_Images}

\newpage
\input{Rebuttal/RS-10-NS}

% \newpage
% \bibliography{bibliography}

\end{document}

%% file: math_commands.tex
\usepackage{amsmath,amsfonts,bm}

\def\eqref#1{equation~\ref{#1}}
\def\1{\bm{1}}

\DeclareMathAlphabet{\mathsfit}{\encodingdefault}{\sfdefault}{m}{sl}
\SetMathAlphabet{\mathsfit}{bold}{\encodingdefault}{\sfdefault}{bx}{n}

%% file: 00-00-definitions.tex
\newcommand{\SceneImage}[1]{{S_{#1}}}
\newcommand{\RetinaEncodingOperator}{\mathrm{\Upsilon}}

\newcommand{\OpticNerveSignal}[1]{{O_{#1}}}
\newcommand{\VisualPercept}[1]{{V_{#1}}}
\newcommand{\VisualPerceptDistorted}[1]{{\tilde{V}_{#1}}}
\newcommand{\PhotoreceptorActivations}[1]{{A_{#1}}}

\newcommand{\PredictedOpticNerveSignal}[1]{{\hat{O}_{#1}}}

\newcommand{\ConeTypesBucket}{\mathrm{C}}
\newcommand{\ConePositionsBucket}{\mathrm{P}}
\newcommand{\LateralInhibitionWeightsBucket}{\mathrm{W}}
\newcommand{\DemosaickerBucket}{\mathrm{D}}
\newcommand{\MotionEstimationBucket}{\mathrm{M}}

\newcommand{\DecodeOperator}{\mathrm{\Phi}}
\newcommand{\EncodeOperator}{\mathrm{\Psi}}
\newcommand{\TranslateOperator}{\mathrm{\Omega}}

\newcommand{\ColorPerceptDim}{N}

\newcommand{\MotionEstimationFunctionTemplate}[2]{\mathrm{\mu}_{\MotionEstimationBucket}(#1,#2)}
\newcommand{\MotionEstimationFunction}{\MotionEstimationFunctionTemplate{\OpticNerveSignal{t}}{\OpticNerveSignal{t+dt}}}

\newcommand{\colorvectorspace}{{$\mathbb{R}^\ColorPerceptDim$}\xspace}
\newcommand{\bucketP}{{$\ConePositionsBucket$}\xspace}
\newcommand{\bucketC}{{$\ConeTypesBucket$}\xspace}
\newcommand{\bucketLI}{{$\LateralInhibitionWeightsBucket$}\xspace}

\newcommand{\bucketDM}{{$\DemosaickerBucket$}\xspace}

\newcommand{\CMFSim}{{$\mathit{CMF\textrm{-}SIM}$}\xspace}
\newcommand{\NScope}{{$\mathit{NS}$}\xspace}

\newcommand{\SPDs}{spectra\xspace}
\newcommand{\SPD}{spectrum\xspace}

%% file: 00-Abstract.tex
\vspace*{-1em}
\begin{abstract}
% The optic nerve signal arriving at the brain is fundamentally different from our experience of vision, and it remains a mystery how the brain decodes color vision purely from a stream of such signals. The brain faces the challenge of disentangling an internal visual perception of the world, with the correct color dimensionality, despite the unknown encoding properties of the eye.
It is a mystery how the brain decodes color vision purely from the optic nerve signals it receives, with a core inferential challenge being how it disentangles internal perception with the correct color dimensionality from the unknown encoding properties of the eye. 
In this paper, we introduce a computational framework for modeling this emergence of human color vision by simulating both the eye and the cortex. Existing research often overlooks how the cortex develops color vision or represents color space internally, assuming that the color dimensionality is known a priori; however, we argue that the visual cortex has the capability and the challenge of inferring the color dimensionality purely from fluctuations in the optic nerve signals. To validate our theory, we introduce a simulation engine for biological eyes based on established vision science and generate optic nerve signals resulting from looking at natural images. Further, we propose a bio-plausible model of cortical learning based on self-supervised prediction of optic nerve signal fluctuations under natural eye motions. We show that this model naturally learns to generate color vision by disentangling retinal invariants from the sensory signals. When the retina contains $N$ types of color photoreceptors, our simulation shows that $N$-dimensional color vision naturally emerges, verified through formal colorimetry. Using this framework, we also present the first simulation work that successfully boosts the color dimensionality, as observed in gene therapy on squirrel monkeys, and demonstrates the possibility of enhancing human color vision from 3D to 4D.
\end{abstract}

%% file: 01-Introduction.tex
\section{Introduction}

% 1. Retinal signals are so different from human color vision
\centerline{\textit{``Color is the place where our brain and the universe meet.'' -- Paul Klee}}

We experience colors in everyday life so effortlessly that it is easy to take the underlying neural computations for granted. In fact, the sensory signals exiting our eye, called optic nerve signals (ONS), are nothing like our color vision (see Fig.~\ref{fig:1} and Supplementary Video~0:30~\footnote{Code, video, dataset, and tutorials are available on our project website: \url{https://matisse.eecs.berkeley.edu}.}). For example, ONS are spatially warped, akin to an image taken with a fish-eye lens, due to varying densities of photoreceptor cells in the retina~\citep{curcio1990human}. ONS does not come in color either -- colors of the scene are spectrally sampled by different types of color sensitive cells (cone cells) in the retina, appearing as a layer of spatial noise in the ONS. Furthermore, other processes, such as lateral inhibition and action potentials, render the image structure barely recognizable, in gradient domain where spatiotemporal ``edges'' dominate. Now, the question is: \textit{how are we still seeing colors?}

% 2
% Indeed, it remains a scientific mystery how the brain decodes color vision from the optic nerve signals it receives. For instance, the brain does not know the number of cone types. It just happens that humans with 3 types of cone cells generally result in 3D ``normal'' color vision, while 2 types result in 2D ``color-blind'' vision~\cite{maxwell1856theory}. \citet{mancuso2009gene} performed gene therapy on color-blind squirrel monkeys that added a third cone type on the retina, revealing that color dimensionality couild be boosted from 2D to 3D, even in adulthood, and found that the transformation was seemingly immediate. How is such a fundamental change in adult color vision even possible? The computational complexity of vision emerging from the ONS runs deep. The brain does not know the foveated layout of cells that spatially warps the ONS, or the lateral inhibition weights of the retina that encode the signals into a quasi-gradient domain. In this paper, we attempt to deconstruct this mystery by proposing a simple, yet complete mechanism for how the brain disentangles clear, high-resolution color vision from the unknown, complex, fine-grained encoding properties of the eye, purely from the optic nerve signals.

% 3
Specifically, this paper introduces a novel computational framework for modeling the emergence of human color vision by simulating the eye and the cortex. 
For the eye, we present a biophysically accurate implementation of a textbook scientific model of the retinal neural circuitry.
% For the eye, we present a flexible simulation engine that can generate biological optic nerve signals for diverse color vision genotypes.
For the brain, we hypothesize a low-level, self-supervised learning mechanism in the cortex that operates purely on the optic nerve signal stream.
%and aims to predict the cellular-level fluctuations in optic nerve signals that occur during small eye movements. We show that the posited self-supervised learning results in emergence of color vision with the correct dimensionality.
For color representation in the brain, we propose modeling color in the brain as a high-dimensional vector, rather than assuming any specific color dimensionality and show that the correct color dimensionality emerges naturally through the proposed learning.

% 4
For eye simulation, our goal is to create a computational engine that takes in any spectral image of the world and outputs the corresponding optic nerve signal stream, based on established vision science. Our model captures how scene form, color, and motion become spatiotemporally entangled in the optic nerve signal stream. 
Our simulation is based on the ``textbook'' model of vision science~\citep{rodieck1998first} for midget, private-line visual pathways, detailed in Section~\ref{sec:method_eye}, and comprising fixational eye motion, spectral encoding by cone cells, foveation, and lateral inhibition.
%which govern the high-acuity color vision near the human fovea. Specifically, we model: natural eye motion, called fixational eye drift;  spectral response functions and cell layout statistics for long, medium and short-wavelength sensitive (L, M, S) cone cells;spatial variation in cell density that is the basis of foveated vision~\cite{curcio1990human}; and lateral inhibition that encodes photoreceptor activations into the gradient domain, increasing the power efficiency of transmission down the optic nerve~\cite{barlow1961possible}.

% \begin{figure*}[t!]
%     \centering
%     \includegraphics[width=\textwidth]{assets/fig1_0829.pdf}
%     \caption{Overview of our proposed framework for modeling the emergence of human color vision. Our simulation engine of biological eyes converts a scene stimulus to a stream of optic nerve signals (Section~\ref{sec:method_eye} \& Video~0:13). We simulate cortical learning purely from these optic nerve signals  (Section~\ref{sec:method_learning}) and show the emergence of color vision. We show how to analyze the emegent neural color qualitatively with Neural Scope (\NScope) and quantitatively with Color Matching Function test Simulator (\CMFSim) (Section~\ref{sec:method_representation}).}
%     \label{fig:1}
% \end{figure*}

\begin{figure*}[t!]
    \centering
    \includegraphics[width=\textwidth]{assets/fig1_0214.pdf}
    \caption{Overview of our proposed framework for modeling the emergence of human color vision. Our simulation engine of biological eyes converts a scene stimulus (hyperspectral image) to a stream of optic nerve signals (Section~\ref{sec:method_eye} \& Video~0:30). We simulate cortical learning purely from these optic nerve signals  (Section~\ref{sec:method_learning}) and show the emergence of color vision. We show how to analyze the emergent neural color quantitatively with Color Matching Function test Simulator (\CMFSim) and qualitatively with Neural Scope (\NScope) (Section~\ref{sec:method_representation}).}
    \label{fig:1}
    \vspace{-10pt}
\end{figure*}

% 5
For brain simulation, we hypothesize that the cortex could disentangle the optic nerve signal from the invariant retinal properties to generate color vision – purely through a self-supervised learning process that aims to predict the constant fluctuations in cellular-level activations of the optic nerve signal during small eye movements.  The neural conditions for such self-supervised learning are biologically plausible in the sense that the cortex continuously receives optic nerve signals under the tiny gaze movements of fixational eye drift.%~\citep{rucci2015unsteady,rucci2018temporal,young2021emulated}.
The theoretical intuition for why such learning might succeed is that eye motion repeatedly draws a static scene image across the invariant spectral and spatial sampling properties of the retina, potentially enabling the retinal properties and scene images to be mutually filtered out of the optic nerve signal where they are entangled.  We show that this simple learning mechanism succeeds at discovering color vision with the correct dimensionality.
% 
% Remarkably, this simple learning mechanism succeeds at discovering clear vision with the correct color dimensionality, as summarized below.

% 6
But first, a somewhat esoteric yet technically critical feature of the modeling framework needs discussion: color representation in the brain. 
Existing research often overlooks how the cortex develops color vision or represents color space internally, assuming that the color dimensionality is known a priori, e.g. 
RGB.
We argue that the cortex has the capability and the task of inferring color dimensionality, purely from fluctuations in the optic nerve signals. Therefore, we propose representing color in the cortex as a high-dimensional vector in $\mathbb{R}^N$ and find that the correct color space and color dimensionality emerge naturally as geometric properties of the hypothesized learning mechanism. We show how to formally quantify and visualize the emergent color space (Section~\ref{sec:method_representation}).

% This representation allows the cortex to possibly encapsulate color space larger than the $3$ dimensions, consistent with the study of functional human tetrachromatic vision~\cite{jordan2010dimensionality}. In exchange, the cortex faces the challenge of constructing a color manifold of unknown dimension $K$ from the sensory signals it receives through the optic nerve. Further, the high-dimensional nature of this representation makes the assessment of internal color dimensionality a non-trivial task, necessitating a formal colorimetric evaluation of the cortical model's color space. To this end, we introduce two measurement methods; formal \CMFSim based on the classical color matching function tests of Maxwell and Grassman~\cite{maxwell1856theory,grassmann1853theorie}, and intuitive \NScope to visualize the learned color space, which together provide a comprehensive evaluation of the model's color dimensionality.

% 7
Remarkably, the hypothesized learning mechanism results in clear color vision of the correct color dimensionality. When the retina contains 1, 2, 3 or 4 types of color photoreceptors (cones), correct color dimensionality emerges: respectively, 1D mono-, 2D di-, 3D tri- or 4D tetra-chromatic vision. In fact, it is an esoteric but well-known fact in vision science that the three types of 2D dichromacy result in highly specific color-spaces (blue-yellow, blue-orange, and teal-pink), which we see emerge naturally.
% 
% Tetrachromacy enjoyed by some species, e.g. goldfish, chickens, tortoises~\cite{jacobs2018photopigments} and an active area of research in women~\cite{jordan2010dimensionality,leeTetrachromacy2024}. 
% 
Even more, the simulation presents a model of how color dimensionality boosting occurred in the squirrel monkeys~\citep{mancuso2009gene}. We simulate injection of the third cone pigment virus into the dichromat retina, which results in boosting from dichromacy to trichromacy.
Intriguingly, the model also shows the possibility that normal human trichromacy could be boosted to tetrachromacy.

In sum, our proposed framework formulates the emergence of human color vision in a computational manner. The simulation engine of realistic optic nerve signals generates training data to this problem, and an intentional and challenging constraint is that the cortical model must strictly learn only from the optic nerve signal with no auxiliary information. This paper presents the first, simple and yet complete existence proof that such cortical inference is possible, and we show that this learning simulation is consistent with various surprising and unexplained vision science phenomena.

%% file: 02-Related_Work.tex
\section{Related Work}

{ % start group for locally modifying gap above and below subsection headings. 

% % Temporarily redefine \subsection to include custom spacing
%   \let\originalsubsection\subsection % Save the original \subsection
%   \renewcommand{\subsection}[1]{%
%     \vspace*{-2pt}% Adjust space before the subsection
%     \originalsubsection{#1}% Call the original \subsection
%     \vspace*{-3pt}% Adjust space after the subsection
%   }

\subsection{Vision Science on Optic Nerve Signal Encoding}
% \subsection{Retinal Functions and Existing Recordings}
In the ``textbook'' model, the retina transforms light into electrical signals through three primary functions:  color sampling via cone cell spectral response functions, lateral inhibition via horizontal neural connections, and spatial sampling via cell positioning (Fig.~\ref{fig:2}). Under daylight, this process begins when light activates cone cells, which are of three types for most humans~\citep{young1801ii,young1802ii,von1867handbuch}, each sensitive to different wavelengths~\citep{stockman2000spectral}. These signals are then modulated by horizontal cells that enhance visual contrast through lateral inhibition~\citep{rodieck1965quantitative,dacey1996horizontal,verweij2003surround}. The signals continue to bipolar and retinal ganglion cells (RGCs), with a direct connection in the fovea via the midget private-line pathway~\citep{dowling1966organization,mcmahon2000fine,wool2018nonselective}, vital for high-resolution color vision. Notably, the cone cell density varies, reaching its peak in the fovea~\citep{osterberg1935topography,curcio1990human}. The axons of the ganglion cells bundle to form optic nerve signals, maintaining their spatial arrangement, which results in a retinotopy in the visual cortex~\citep{holmes1918disturbances,tootell1982deoxyglucose,dougherty2003visual}. Fixational eye movements cause dynamic photoreceptor activation at all times by constantly shifting the gaze, even when focusing on static objects~\citep{rucci2015unsteady,young2021emulated,martinez2013impact}.

Recordings of real optic nerve signals exist, but cannot be used for our cortical simulations because they are orders of magnitude too low resolution (from only a few thousand cells) and only in response to grayscale rather than color imagery~\citep{litke2004does,brackbill2020reconstruction,marre2017multi,liu2022simple}. 
% There is a study that documented RGC responses to color stimuli~\cite{schottdorf2021quantitative}, but it only captured responses from $47$ RGCs, and these responses did not match anywhere close to the number of cones in human eye (i.e. $6$ million~\cite{curcio1990human}).
To overcome these data limitations, we present a simulation engine for the midget private-line pathway in the fovea in Section~\ref{sec:method_eye}. 

% % datasets
% \cite{berry1997structure} 137 cells from 10 retinas in salamander + 45 cells from 2 retinas in rabbit.
% \cite{schneidman2006weak} 40 cells from salamandar
% \cite{marre2017multi} 160 salamander retinal ganglion cells
% \cite{liu2022simple} 156 RGC recordings from 9 retinas of salamander and mouse.
% \cite{sibille2022high} 1,199 RGC axons from 27 recordings of 24 mice.
% \cite{reinhard2021visual} human 342 ganglion cells
% \cite{kim2021nonlinear} >1000 human RGCs
% \cite{litke2004does} multi-electrode recordings from the peripheral macaque retina 

\subsection{Color Representation in Computational Neuroscience}
It is an open question how to meaningfully model neural representations of color in simulations of visual perception. 
Most computational neuroscience sidesteps this issue, ``hard-coding'' a dimensionality of $3$, representing and constraining cortical color to tristimulus values, such as RGB~\citep{parthasarathy2017neural,botella2018nonlinear,brackbill2020reconstruction,kim2021nonlinear,zhang2022image,wu2022maximum}, LMS~\citep{young1802ii,von1867handbuch} or cone-opponent space~\citep{derrington1984chromatic,macleod1979chromaticity}. 
Constraining cortical models to such a ceiling of three for dimensionality clearly conflicts with the study of a functional tetrachromat observer with 4D color vision~\citep{jordan2010dimensionality,rezeanu2021explaining}. Instead, we model the brain with the capability and the challenge of deducing the inherent color dimensionality encoded in the optic nerve signals.
% 
%Therefore, when establishing the representation of cortical color, it is important to refrain from making any overt assumptions about its dimensionality.

\subsection{Theory on How Vision Emerges in the Brain}
% self-supervised learning
In computational vision science modeling, it is often overlooked that the cortex relies solely on a stream of optic nerve signals to discover color vision, with no access to a teacher signal or perceptual ground truth.
Rather, various efforts have been made to reconstruct visual stimuli from neural responses by giving the cortical model access to ground truth stimulus image~\citep{naselaris2009bayesian,nishimoto2011reconstructing,parthasarathy2017neural,botella2018nonlinear,brackbill2020reconstruction,kim2021nonlinear}, but the neural reality is that the brain never has direct access to the visual scene.
This characteristic makes the human visual system a quintessential example of self-supervised learning, which is a growing field in computer vision~\citep{de1993learning,chen2020simple,he2022masked}. In this work, we propose a learning principle in the cortex which aims to predict the cellular-level fluctuations in activation that occur during small eye movements, which is associated with the idea of temporal prediction~\citep{palmer2015predictive, lotter2016deep, singer2018sensory, singer2023hierarchical}, as well as the broader concept of predictive coding~\citep{rao1999predictive,srinivasan1982predictive}. It is also closely linked to the sensorimotor contingency theory from cognitive science~\citep{o2001sensorimotor} and slow feature analysis~\citep{hinton1990connectionist,foldiak1991learning} which suggests that the brain learns to anticipate the sensory outcomes of motor actions (e.g. eye movements) and filters out invariant (slow) features.

\subsection{Theory on How Brains Infer Color Dimensionality}
% equivariance, inference of retinal invariants
Previous studies has investigated inference of invariant retinal properties from sensory signals. For instance, research has demonstrated the ability to deduce cone cell types~\citep{wachtler2007cone,brainard2008trichromatic,benson2014unsupervised} and the positions of photoreceptors~\citep{maloney_ahumada,brainard2008trichromatic} via statistical methods from sensory signals. \cite{brainard2008trichromatic} hinted at analyzing sensory inputs from different time points during fixational drift as a way to reveal retinal features, and in this paper we provide a computational realization of this idea with a specific learning mechanism that achieves complete disentanglement of retinal invariants from optic nerve signals.

% - complexity of human visual system
%     - it is really reasonable to say that cd is not all of color vision.
%     - this paper focuses on that.
%     - there are a lot more to color, this paper is not going to do that.
%         - adaptation
%         - watercolor illusions
%     - 1 sentence that contains 5-10 citations.

% \subsection{\red{Measurement of Color Dimensionality in Human Color Perception}}
% \red{The Young-Helmholtz trichromatic theory~\citep{young1802ii, von1867handbuch} established that the 3-dimensional nature of human color vision is determined by the number of cone types, forming the basis of modern color reproduction. This was confirmed by Maxwell and Grassmann’s color-matching experiments~\citep{maxwell1856theory, grassmann1853theorie}, but Jacobs's recent work~\citep{jacobs2018photopigments} suggests that color dimensionality is much more nuanced that the number of cone types, illustrating complex interactions within the visual system, extending beyond the properties of cones alone.}

\subsection{Measurement of Color Dimensionality in Human Color Perception}

In this paper, we need to rigorously measure the color dimensionality of the emergent color from cortical simulation.  To do so, we adapt Maxwell's famous color-matching experiments~\citep{maxwell1856theory}, which laid the foundation of colorimetry that remains at the heart of all color reproduction technology today. Color matching experiments confirmed the trichromatic theory~\citep{young1802ii,von1867handbuch,grassmann1853theorie} in which the 3-dimensional nature of human color vision has its basis in the three different cone types in the human retina. Jacobs's recent review~\citep{jacobs2018photopigments} of color dimensionality in animal vision, however, reminds the community that the dimensionality of color vision does not automatically equal the number of cone types, and that the most rigorous way to measure it remains Maxwellian color matching -- as we follow in this work.

% \subsection{\red{Complexity of Human Color Vision}}
% \red{Color dimensionality represents a fundamental aspect of human vision, but the parvocellular pathway and neural processes spanning V1, V2, and V4 add significant complexity~\citep{livingstone1987psychophysical, zeki1978functional, li2014perceptual, liu2020hierarchical, angueyra2022predicting}. Models like CIELAB~\citep{CIE1976} and Color Appearance Models~\citep{fairchild2013color} incorporate non-linear photoreceptor responses~\citep{krauskopf1992color, angueyra2022predicting}, as well as adaptation to illuminants and contextual effects~\citep{von1902theoretische,land1977retinex,FOSTER2011674}. This paper focuses on dimensionality while leaving topics like adaptation and higher-order visual effects for future work.}

\subsection{Complexity of Human Color Vision}
This paper focuses on color dimensionality because it is the foundational characteristic of an observer's color experience, but many layers of further perceptual complexity exist atop that foundation. Examples include chromatic adaptation (color constancy)~\citep{von1902theoretische,land1977retinex,FOSTER2011674}, perceptual nonuniformity across colorspace~\citep{CIE1976,macadam1942visual}, complex contextual interactions~\citep{fairchild2013color}, and even surprising flood-fill features~\citep{pinna1987effetto,pinna2001surface}. Parts of these perceptual phenomena are scientifically mapped to neural correlates, such as nonlinear photoreceptor responses~\citep{krauskopf1992color, angueyra2022predicting}, or parvocellular pathway and neural processes spanning V1, V2, and V4~\citep{livingstone1987psychophysical, zeki1978functional, li2014perceptual, liu2020hierarchical, angueyra2022predicting}. This paper leaves these additional layers of perceptual complexity as future modeling work.

%% file: 03-Simulation_Engine.tex
% \section{Methods}

\begin{figure*}[t!]
    \centering
    \includegraphics[width=0.99\textwidth]{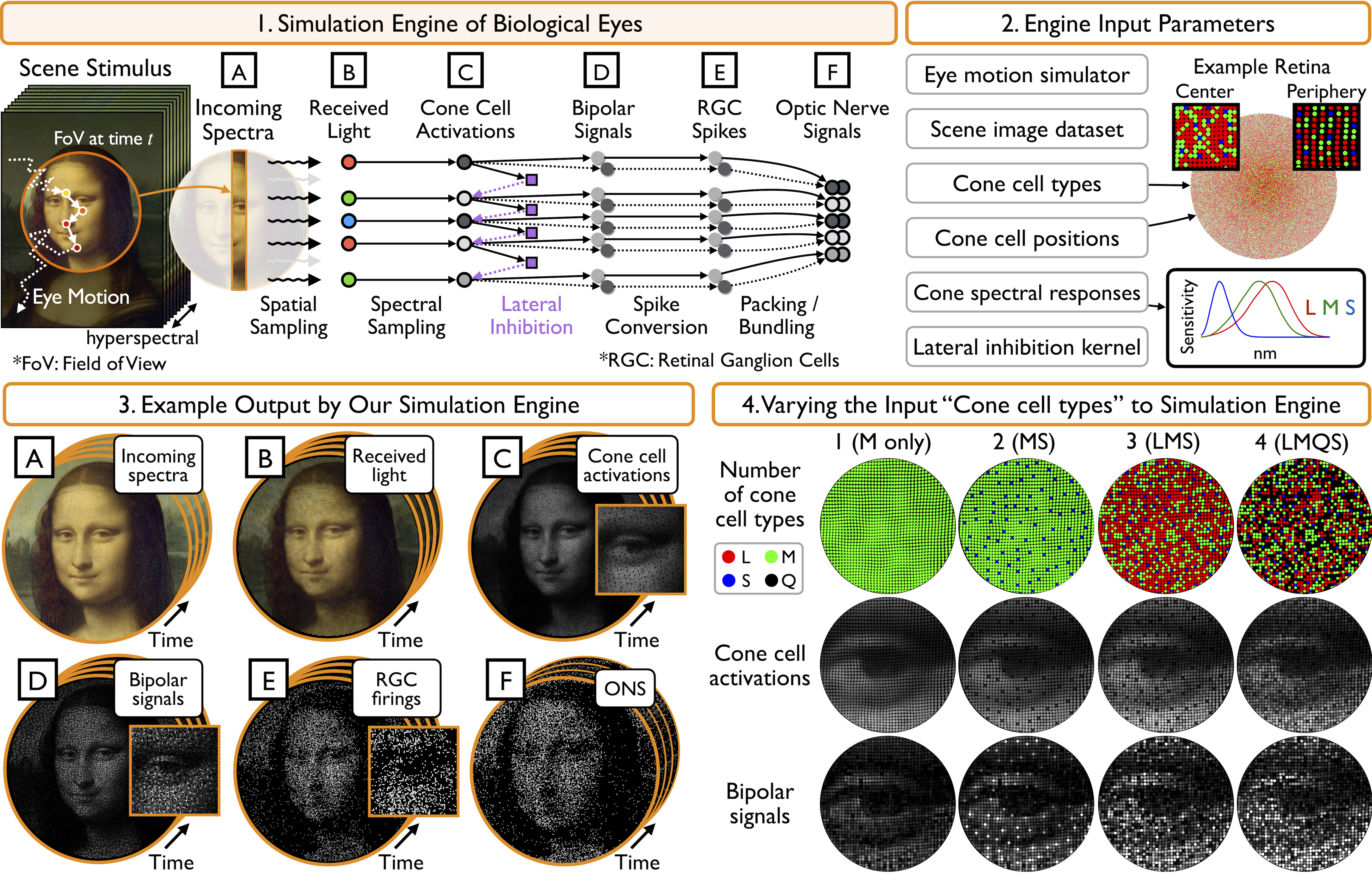}
    \caption{Overview of our simulation engine of biological eyes. 1. This engine takes a scene stimulus as an input, processes it through a ``textbook'' model of eye motion and retinal neural circuitry, to generate a stream of optic nerve signals. 2. This engine accepts custom eye and retina parameters. 3. It allows visualization of neural signals in steps (A-F), illustrating the progressive entanglement of scene imagery with retinal properties. 4. Visualization of changing one of the input parameters, the number of cone cell types -- showing that the signals become noisier as the number increases.}
    % \vspace{-pt}
    \label{fig:2}
\end{figure*}

\section{Simulation Engine for Biological Eyes and Optic Nerve Signals}\label{sec:method_eye}

We model the primary functions of the human retina based on the known science.
% The scene imagery is sampled from a dataset of hyperspectral images~\citep{arad2022ntire} (Fig.~\ref{fig:2}.A). 
The inputs to our retina model are hyperspectral images (Fig.\ref{fig:2}.A; details in Appendix~\ref{sec:methods-scene-images}).
A model of fixational eye drift~\citep{rucci2015unsteady,young2021emulated,martinez2013impact} generates a sequence of frames that sample different parts of the image. 
%\red{(the Supplementary Video also visualizes saccades for illustrative purposes)}. 
Each frame is projected on the retina, stimulating a randomized array of cone cells, according to known cell density variation as a function of eccentricity (distance from the fovea)~\citep{curcio1990human} and known statistics of different cone cell types~\citep{carroll2000flicker} (Fig.~\ref{fig:2}.B). Each cone cell converts the scene light into photoreceptor activations based on cone type spectral response functions~\citep{stockman_spectral_1999,stockman2000spectral} (Fig.~\ref{fig:2}.C), followed by lateral inhibition from horizontal cells~\citep{wool2018nonselective,rodieck1965quantitative} (Fig.~\ref{fig:2}.D). These signals are transformed into spike trains (Fig.~\ref{fig:2}.E), then bundled into optic nerve signals (ONS), resulting in spatial distortion (Fig.~\ref{fig:2}.F), turning a color scene stimulus into a noisy, spatially distorted ONS. Additional details are in Appendix~\ref{sec:supp-retina} and Video 1:35.

One observation to make is that there are three invariant properties of the simulated retina (i.e. retinal invariants): cell positions, cell types, and horizontal cell connections. These properties are held constant during generation of ONS for a particular eye, but we use the engine with different values for these properties to generate ONS datasets for a diverse set of eyes. For example, we create datasets with different cone types, including monochromatic (L, M, S), dichromatic (LM, LS, MS), trichromatic (LMS) and tetrachromatic (LMSQ) configurations (Fig.~\ref{fig:2}.4). The generated ONS becomes spatially noisier as the number of cone types increases, and we study how the cortex infers the inherent color dimensionality of these eyes purely from the differences in their respective ONS. 
% We will release the dataset and code for generating ONS upon acceptance.

%% file: 04-Cortex.tex
\begin{figure*}[t!]
    \centering
    \includegraphics[width=0.99\textwidth]{assets/fig3_0829.pdf}
    \caption{Overview of our hypothesized cortical learning mechanism and exclusive study of the learning behavior of the cortical model with trichromatic retina. 1. Given the stream of optic nerve signals as the only input data, the cortical model aims to predict the next ONS from the current one with 3 learnable functions, decoder $\Phi$, translation operator $\Omega$ and re-encoder $\Psi$. 2. Prediction error decreases as learning progresses, converging after 100K learning steps. 
    3. During learning, the color dimensionality of the internal percepts transition from 1D, 2D to 3D, formally measured by \CMFSim and visualized by \NScope~(Fig.~\ref{fig:4} \& Section~\ref{sec:method_representation}). 
    4. The cortical model infers the retinal invariant properties during learning: cell positions \bucketP (higher density in fovea), cone cell types \bucketC, and lateral inhibition weights \bucketLI (center-surround receptive field).}
    \label{fig:3}
\end{figure*}

\section{Simulating Cortical Learning and Emergence of Color Vision}\label{sec:method_learning}

Our cortical model is structured to learn three functions in a pipeline (Fig.~\ref{fig:3}.1): 1. $\DecodeOperator$ that decodes the optic nerve signal at time $t$ into its internal percept; 2. $\TranslateOperator$ that translates this percept according to eye motions inferred from the signal over a short time interval, $dt$; and 3. $\EncodeOperator$ that re-encodes the translated percept back into a predicted optic nerve signal at time $t+dt$, which is compared against the real signal received at that time. Mathematically, given the optic nerve signal $\OpticNerveSignal{t}$ at time $t$; 
\begin{equation}
    \PredictedOpticNerveSignal{t+dt} = \EncodeOperator(\TranslateOperator( \DecodeOperator(\OpticNerveSignal{t})))
\end{equation}
where $\PredictedOpticNerveSignal{t+dt}$ is the predicted ONS at time $t+dt$.
Here the task of decoder $\DecodeOperator$ is analogous to inverting the retinal processes to transform $\OpticNerveSignal{t}$ into the visual percept image $\VisualPercept{t}$ (i.e. $\VisualPercept{t} \leftarrow \DecodeOperator(\OpticNerveSignal{t})$). Likewise, the re-encoder function $\EncodeOperator$ resembles the retinal processes, as it aims to reproduce an optic nerve signal from the visual percept (i.e. $\OpticNerveSignal{t} \leftarrow \EncodeOperator(\VisualPercept{t})$). Therefore, $\DecodeOperator$ and $\EncodeOperator$ are pseudo-inverses. 

The learning objective is to minimize the prediction error $E_\mathrm{prediction}$, the difference between predicted and real optic nerve images at time $t+dt$, such that:
\begin{equation}
    E_\mathrm{prediction} = \| \OpticNerveSignal{t+dt} - \PredictedOpticNerveSignal{t+dt} \|_2^2 
    = \| \OpticNerveSignal{t+dt} - \EncodeOperator(\TranslateOperator( \DecodeOperator(\OpticNerveSignal{t}))) \|_2^2   
\end{equation}
where $\OpticNerveSignal{t+dt}$ is the real observed ONS at time $t+dt$.
% 

% \subsection{Inference of Retinal Invariant Properties}

% Three main ideas drove the evolution of our selected model features. First, we reasoned that it would be a big step towards successful decoding and re-encoding if the cortex could infer the key encoding properties of each cone cell.  

Three main ideas drove the evolution of our selected model features. First, we reasoned that it would be a big step towards successful decoding and re-encoding if the cortex could infer the key encoding properties of each cone cell.  We gave the cortical model sets of learnable parameters, which we call ``neural buckets'', in which to store and update guesses of these properties during learning. The buckets contain the following information for each cone cell: 2D position in visual space (\bucketP), cone spectral identity (\bucketC), and lateral inhibition weights (\bucketLI) to neighboring cones.

The second main idea was the observation that decoder $\DecodeOperator$ and encoder $\EncodeOperator$ are mathematically factorizable into a pipeline of subfunctions, such that:
\begin{align}
    \DecodeOperator &= \DecodeOperator_\ConePositionsBucket \circ \DecodeOperator_\ConeTypesBucket \circ \DecodeOperator_\LateralInhibitionWeightsBucket \nonumber\\
    \EncodeOperator &= \EncodeOperator_\LateralInhibitionWeightsBucket \circ \EncodeOperator_\ConeTypesBucket \circ \EncodeOperator_\ConePositionsBucket. \nonumber
\end{align}
Each sub-function is an operator conditioned on its corresponding neural bucket, cell positions $\ConePositionsBucket$, cone spectral types $\ConeTypesBucket$, and lateral inhibition weights to neighboring cells $\LateralInhibitionWeightsBucket$. 
In case of decoder $\DecodeOperator$, it first executes inversion of lateral inhibition using $\DecodeOperator_\LateralInhibitionWeightsBucket$ in order to estimate the activations of each photoreceptor associated with an optic nerve axon; second, projects scalar cone activations into \colorvectorspace by $\DecodeOperator_\ConeTypesBucket$ using inferred spectral identities in \bucketC (and interpolating color across space -- the third main idea below); and finally inverting the spatial distortion of foveation via $\DecodeOperator_\ConePositionsBucket$. The re-encoding function is a pipeline of analogous subfunctions in reverse order: re-applying spatial warping with $\EncodeOperator_\ConePositionsBucket$; re-projecting color into scalar photoreceptor activations with $\EncodeOperator_\ConeTypesBucket$; and re-applying lateral inhibition with $\EncodeOperator_\LateralInhibitionWeightsBucket$. Further implementation details are provided in Appendix~\ref{sec:supp-cortex}.

The third main idea was that, in order to accurately re-encode after translation, the model needs to learn to interpolate color information spatially, because there is only one cone type at each point on the retina. 
This need to interpolate is analogous to the situation in cameras with image sensors~\citep{bayer1976color,fossum1997cmos,kimmel1999demosaicing} that physically sample only one of the R,G,B channels at each pixel, and fundamentally require demosaicking algorithms to interpolate full R,G,B values at all pixels. The required interpolation function in the cortical model is more complex because the spectral sampling pattern is random, so learning this function is entangled with correctly resolving \bucketC. We enable and force the cortical model to learn the color interpolation function by representing it as a convolutional neural network with neural bucket parameters \bucketDM (Appendix~\ref{sec:methods-cortex-C}).

An important detail is the handling of eye motion in the simulation. In one experiment we show that the model can learn a subfunction that estimates the eye motion translation between times $t$ and $t+dt$ purely from the optic nerve signals at those times. 
The main computational challenge here is the spatial warp in optic nerve signals. The uniform translation in stimulus space corresponds to a non-uniform translation in the optic nerve signal space, which makes the prediction of the eye motion dependent on the inference of cell positions \bucketP. We find that the cortical model iteratively updates the neural bucket \bucketP from imperfect initial eye motion estimate, which helps to improve the prediction of eye motion, and vice versa. This inference converges to the correct eye motion estimate, as the model learns to minimize prediction error (Appendix~\ref{sec:methods-cortex-M}).

With this pipeline of subfunctions and associated neural buckets, the hypothesized learning is equivalent to parallel numerical optimization of all neural buckets in striving to minimize prediction error. We simulate learning using stochastic gradient descent~\citep{kingma2014adam}.

%% file: 05-Representation.tex
% \section{Neural Representation of Color Space and Its Dimensionality}\label{sec:method_representation}
\section{Neural Representation of Color Space and Analysis of Emergent Color Dimensionality}\label{sec:method_representation}

% \begin{figure*}[t!]
%     \centering
%     \includegraphics[width=\textwidth]{assets/fig4_0827.pdf}
%     \caption{Overview of \CMFSim. 1. The goal of \CMFSim is to find the minimum number of primary colors that can be matched to any target colors. A match color is a weighted sum of primary colors, and coefficients are iteratively updated to minimize the perceptual error until convergence. 2. Example output for the cortical model with trichromat retina (3 cone cell types). The matching errors for all target spectral light, ranging from 400nm to 700nm, first converge to zero error with 3 primaries, formally indicating that the model has 3D color vision.}
%     \label{fig:4}
% \end{figure*}
\begin{figure*}[t!]
    \centering
    \includegraphics[width=\textwidth]{assets/fig4_1126.pdf}
    \caption{Overview of our measurement methods of emergent color dimensionality, Color Matching Function Test Simulator (\CMFSim) \& Neural Scope (\NScope). 
    % 1. The goal of \CMFSim is to find the minimum number of primary colors that can be matched to any target colors. A match color is a weighted sum of primary colors, and coefficients are iteratively updated to minimize the perceptual error until convergence. 2. \NScope is a visualization tool for visual percepts, independent from the cortical learning loop. It is parameterized as a learnable $N\times 3$ matrix, and optimized to minimize the projection error to the target RGB image. 3. Example output of \CMFSim for the cortical model trained with trichromat retina after convergence. The matching errors for all target spectral light, ranging from 400nm to 700nm, first converge to zero error with 3 primaries, formally indicating that the model has 3D color vision.
    1. \CMFSim determines the minimum number of primary colors needed to match any target color by iteratively updating coefficients to minimize perceptual error. 2. \NScope visualizes visual percepts independently of the cortical learning loop, optimized as a learnable $N\times 3$ matrix to minimize projection error to the target RGB image. 3. Example \CMFSim output for a trichromat retina-trained cortical model shows matching errors for 400–700nm spectral light converging to zero with three primaries, confirming 3D color vision.
    }
    \label{fig:4}
    % \vspace{-5pt}
\end{figure*}

We model color in the cortex as a vector in high-dimensional space, \colorvectorspace. This decision represents our view that the brain has both the freedom and the challenge of somehow inferring the intrinsic color dimensionality of the visual signals it is receiving along the optic nerve. 
Specifically, we define a cortical decoding function $\DecodeOperator$ that takes the optic nerve signal $\OpticNerveSignal{t}$ at time $t$ and transforms it into its internal visual percept $\VisualPercept{t}$ (Fig.~\ref{fig:3}), such that:
\begin{equation}
    \VisualPercept{t} = \DecodeOperator(\OpticNerveSignal{t})
\end{equation}
where each pixel in $\VisualPercept{t}$ is a $N$-dimensional vector. Inside \colorvectorspace, we assume that cortical color space emerges as a $K$-dimensional manifold. To measure this intrinsic dimensionality $K$, we introduce two methods: formal, numerical \CMFSim; and intuitive, visual \NScope.
% \red{Color Matching Function test Simulator} (\CMFSim); and intuitive, visual \red{Neural Scope} (\NScope).

\CMFSim (Color Matching Function Test Simulator) is our tool to formally quantify the color dimensionality that has emerged in the $\mathbb{R}^N$ color space of the cortical model (Fig.~\ref{fig:4}.1). \CMFSim treats the retina model and cortical model as a black-box color observer.
Specifically, we limit ourselves to showing the model two patches of scene color at random, distinct locations on the retina to reflect the effect of real-world viewing conditions, obtaining only a scalar score as feedback to indicate the difference in color appearance betweem the two patches (with zero representing a color match).
This interface is intentionally limited and identical to color matching experiments with human subjects. Then, resting on the formal, technical bedrock for colorimetry established by \citet{maxwell1856theory} and \citet{grassmann1853theorie}, we exhaustively probe to determine the minimum number of color primaries needed to match any test color through linear combination (Fig.~\ref{fig:4}.1 \& Appendix~\ref{sec:supp-cmfsim}). For example, Maxwell found that most humans require $3$ primaries and are formally trichromatic, but red-green colorblind persons require only $2$ primaries and are dichromatic. We model diverse color observers and measure the color dimensionality of their emergent vision.

\NScope (Neural-Scope) is our tool to display emergent colors in $\mathbb{R}^N$ by projecting them into RGB color space (Fig.~\ref{fig:4}.2).
\NScope is a learnable $N\times 3$ matrix mapping the emergent cortical color manifold in $\mathbb{R}^N$ linearly to conventional RGB color, enabling visual inspection of emergent color vision.
To compute \NScope, we first project our hyperspectral image dataset in two ways: (1) using the retina and learned cortical model to map hyperspectral images into the $\mathbb{R}^N$ cortical colorspace, and (2) using conventional color processing to convert hyperspectral data to RGB via CIEXYZ~\citep{commission1931commission}. \NScope is then determined as the least squares transform from the former data in $\mathbb{R}^N$ to the latter in RGB.
Importantly, \NScope is independently optimized as a visualization parameter, separate from the main cortical loop, ensuring that the cortex processes ONS without any exposure to input images in either hyperspectral or RGB form.
\NScope provides striking color visualizations and complementary intuition visually, which are fully consistent with formal \CMFSim results.
For example, \NScope allows us to compute and contrast the color palettes of the three different types of human dichromacy (Fig.~\ref{fig:5}). 

In sum, we model cortical color as vectors in \colorvectorspace, allow color space to emerge naturally through the hypothesized learning mechanism, and measure the emergent color space's dimensionality quantitatively with \CMFSim and analyze it visually with \NScope.

%% file: 06-Results.tex
\section{Simulation Results - Emergence of Color Vision}\label{sec:result_emergence}

% Figure~\ref{fig:3}.2 begins with a model of typical human retina containing L, M and S cones, and illustrates the time-varying behavior of the cortical model as it learns color vision. The visualized prediction error decreases as the training progresses, and the internal percept converges to $3$-D color vision both formally (\CMFSim) and visually (\NScope imagery), shown in Figure~\ref{fig:3}.3. Notably, the visual timeline highlights that the cortical model learns color vision one dimension at a time: achieving monochromacy at 700 learning steps, dichromacy at 1,200 steps, and converging to trichromacy after $10^6$ steps. 
% % This trend is consistent with child behavioral psychology findings~\cite{brown1990development}, illustrating the theoretical possibility that immature color vision in human infants is partially caused by the cortical learning.
% At convergence, the cortical model has accurately inferred all retinal properties: spectral identity of each cell, cell positions and lateral inhibition neighbor weights (Figure~\ref{fig:3}.4). 

Figure~\ref{fig:3}.2 begins with a model of typical human retina containing L, M and S cones, and illustrates the time-varying behavior of the cortical model as it learns color vision. The visualized prediction error decreases as the training progresses, and the internal percept converges to $3$-D color vision both formally (\CMFSim) and visually (\NScope imagery), shown in Figure~\ref{fig:3}.3. Notably, the visual timeline highlights that the cortical model learns color vision one dimension at a time: achieving monochromacy at 700 learning steps, dichromacy at 1,200 steps, and converging to trichromacy after $10^6$ steps. 
% This trend is consistent with child behavioral psychology findings~\cite{brown1990development}, illustrating the theoretical possibility that immature color vision in human infants is partially caused by the cortical learning.
At convergence, the cortical model has accurately inferred all retinal properties: spectral identity of each cell, cell positions and lateral inhibition neighbor weights (Fig.~\ref{fig:3}.4).
% 
% \CMFSim results show close alignment with human psychophysical measurements~\citep{stiles1955interim} (Appendix~\ref{sec:supp-SandB}) and consistently exhibit 3D color vision across noise initializations (Appendix~\ref{sec:supp-consistency}).
For a cortical model trained with a trichromat retina, \CMFSim results closely align with human psychophysical data~\citep{stiles1955interim} (Appendix~\ref{sec:supp-SandB}) and the model consistently demonstrates 3D color vision across different noise initializations (Appendix~\ref{sec:supp-consistency}), highlighting our model’s validity and robustness.
% 
% The emergence of 3D color vision is consistent across different noise initializations (Appendix~\ref{sec:supp-consistency}).
% We varied the ratio of L:M cone cells in the simulated retinal patch, testing 2:1 (original), 1:1 (variant 1), and 1:2 (variant 2). In all cases, the configurations converged to 3D color vision, indicating that the framework is adaptable to variations in cone distributions.}

The remainder of this section shows \CMFSim and \NScope results at convergence for a diversity of simulation scenarios. Fig.~\ref{fig:4}.2 dissects the \CMFSim analysis for the case where the retina contains $3$ cone types. The graphs show color matching function results using optimal sets of color primaries from $0$ to $4$ primaries, along with the residual perceptual error as a function of wavelength. 
The area under the curve (AUC) for each error graph suggests that the error falls to near-zero only with at least 3 primaries -- this formally proves that the color dimensionality is $3$. %, as expected for this retina with $3$ spectral types of cone cells. 
% Finally, consistent with this formal \CMFSim result, \ref{fig:5}.1.1 contains \NScope visualization of the internal percept, showing that full trichromat color imagery emerges, including the rainbow spectrum image at the bottom. 

Figure~\ref{fig:5} presents results of the hypothesized cortical learning in a diversity of color observers where the retina contains different numbers of cone cell types. 
This shows that the model learns $K$-dimensional color vision when the retina contains $K$ cone types. That is, when the retina contains 1, 2, 3 or 4 cell types, the cortical model converges on mono, di, tri, or tetrachromat color vision, formally quantified with \CMFSim (further analysis of tetrachromat models in Appendix~\ref{sec:supp-tetrachromacy}). And qualitatively, we observe that the \NScope images are grayscale for $K=1$, colorblind with only shades of blue and yellow for $K=2$, and full color with all trichromatic hues only with $K=3$ (Fig.~\ref{fig:5}.1). 
Use of \NScope is limited  to color dimensionality up to 3, and is to not applicable to $K=4$ case, but \CMFSim formally confirms that 4-dimensional color emerges there. 
All variants of retinas with two cone types (protanopia, deuteranopia and tritanopia), converge to 2D color vision. But more striking, \NScope reveals hue shifts among these models that are consistent with color vision deficiencies studies~\citep{brettel1997computerized} (yellow-blue hues for deuteranopia / protanopia~\citep{judd1948color,graham1959studies} and 
% \TODO{teal-orange} 
teal-pink hues for tritanopia~\citep{alpern1983trit}) (Fig.~\ref{fig:5}.2).

\begin{figure*}[t!]
    \centering
    \includegraphics[width=\textwidth]{assets/fig5_0214.pdf}
    \caption{Results of simulating emergence of color vision from various retinas, with analysis of learned color dimensionality using qualitative visualization (\NScope) and formal methods (\CMFSim). 1. Cortical models trained with dataset generated with retinas containing 1, 2, 3, 4 cone types result in  mono-, di-, tri, tetrachromatic color vision, respectively. 2. Qualitative color of dichromat variants is consistent with known vision science on color vision deficiency.	3. Control experiment with a trichromat retina, but with cortical learning deliberately removed: \CMFSim measures color  as 1-D, highlighting that cortical learning is necessary for emergence of correct color dimensionality.}
    \label{fig:5}
    % \vspace{-20pt}
\end{figure*}

% \red{
% Building on this, we address the theoretical question of whether cortical inference is essential for color vision. To investigate, we compare two scenarios: (1) a control experiment baseline, where optic nerve signals are represented directly as internal percepts without cortical inference, and (2) a learning-based model, where cortical processing is involved in shaping color perception. We demonstrate that the control experiment baseline fails to pass standard color matching tests, as \CMFSim reveals the color dimensionality to be reduced to $1$-D for observers with $3$ cone types (Appendix D). In contrast, the learning-based model (Fig.5) successfully reconstructs full color dimensionality, highlighting the critical role of cortical inference in color perception.
% As a control experiment for Figure~\ref{fig:5}, we address the question of whether cortical inference is essential for color vision. To investigate this, we compare two scenarios: (1) a control baseline, where optic nerve signals are directly represented as internal percepts without cortical inference, and (2) a learning-based model, where cortical processing shapes color perception. The control baseline fails to pass standard color-matching tests, as \CMFSim shows that color dimensionality reduces to $1$-D even for observers with $3$ cone types (Appendix~\ref{sec:supp-baseline}). In contrast, the learning-based model (Fig.~\ref{fig:5}) successfully reconstructs full color dimensionality, underscoring the critical role of cortical inference in achieving accurate color perception.
In a control experiment, we verify that cortical learning is essential for color vision by comparing two scenarios: (1) a baseline where optic nerve signals directly form percepts, with all cortical learning deliberately removed (Fig.~\ref{fig:5}.3), and (2) the proposed model including cortical learning (Fig.~\ref{fig:5}.1, LMS case). The control baseline experiment fails standard color-matching tests, reducing color dimensionality to $1$-D despite $3$ cone types, demonstrating that cortical learning is indeed necessary for emergence of correct color dimensionality (details in Appendix~\ref{sec:supp-baseline}). In contrast, the proposed learning-based model results in correct $3$-D color, as expected.

Figure~\ref{fig:6} simulates boosting of color dimensionality in adulthood by gene therapy. Previous genetic studies~\citep{jacobs2007emergence,mancuso2009gene,zhang2017gene} demonstrated that the introduction of an additional class of photopigments in the mammalian cone mosaic, even in adulthood via gene therapy, resulted in a new dimension of chromatic sensory experience. Here, we simulate the experiment performed in squirrel monkeys~\citep{mancuso2009gene}, finding simulation results consistent with the noted boost in color dimensionality (Fig.~\ref{fig:6}.1). 
First, we model the vision of an adult male squirrel monkey with a protanopic retina (M and S cones only) and a cortical model that has converged. Next, we simulate the effects of gene therapy, by modifying the retina model so that a random subset of cones begin to express L opsin. Immediately after this retinal change, \NScope continues to show blue-yellow dichromatic vision. However, if we allow the cortical model to continue the hypothesized self-supervised learning, vision re-converges to boosted, 3-dimensional color (Fig.~\ref{fig:6}.1). As an aside, the real gene therapy experiment~\citep{mancuso2009gene} resulted in expression of both M and L photopigment in affected cones, but the relative amounts are difficult to ascertain. 
We additionally model scenarios in which equal M and L expression occurs or a variable amount of L relative to M at each cell, and both scenarios converge to trichromacy as well (Appendix~\ref{sec:supp-boosting}). These result indicate that the hypothesized self-supervised learning can explain experimental boosting of color dimensionality in adulthood consistent with \citet{mancuso2009gene}, if the hypothesized learning is assumed to occur continuously even in adulthood. 

Figure~\ref{fig:6}.2 further demonstrates that simulations of boosting color dimensionality succeed  even in 3D$\rightarrow$4D cases by the addition of a fourth cone 
% between M and L peak sensitivities as in naturally-occurring human tetrachromacy (human Q) and even addition of cones with different peak sensitivities 
between S and M cones. This simulation results present the first theoretical work that highlights the possibility of boosting humans with trichromatic vision to tetrachromat one by addition of a fourth photopigment. Further details, including the study of human tetrachromat observer model, are described in Appendix~\ref{sec:supp-tetrachromacy}. 
% 
% We will release all the code and trained weights of our cortical model upon acceptance.

\begin{figure*}[t!]
    \centering
    \includegraphics[width=0.7\textwidth]{assets/fig6_0214.pdf}
    \caption{Simulated experiments for boosting color dimensionality via gene therapy. 1. 2D dichromat model is boosted to 3D trichromacy by mutating some M cones to express L opsins, with cortical learning re-converging to 3D color. 2. 3D trichromat model is boosted to 4D tetrachromacy by adding a fourth cone type between M and L cones, with cortical learning re-converging to 4D color.}
    % \vspace{-10pt}
    \label{fig:6}
\end{figure*}

%% file: 07-Conclusion.tex
\section{Conclusion}

% ideas on how to improve each pillar
In this work, we presented a framework for modeling the emergence of color vision in the human brain. We introduced a simulation engine for optic nerve signals, simulated cortical learning purely from such input signals, and measured the emergent color dimensionality qualitatively with \NScope and quantitatively with \CMFSim.

We believe that the critical contribution of this work actually lies in the computational formulation of the problem in human perception itself -- given access to only a stream of optic nerve signals, we probe how to meaningfully model the emergence of color vision in the cortex via simulated learning.  Once the problem is formulated this way, it may not be entirely surprising to machine learning researchers that color with the correct dimensionality can be inferred; however, from the vision science perspective, this new approach is in many ways a foreign way to formulate the problem, because it is more common to think color comes from hardwired neural circuits (Appendix~\ref{sec:supp-baseline}).

The connection to camera imaging systems is noteworthy as well. The proposed computational framework is akin to a camera processing pipeline attached to an unknown, random color filter array pattern, where even the number of different color filters is a mystery. This is related to a branch of computational imaging called ``auto calibration'' that jointly solves for the scene and unknown system calibration parameters from measurements.  One could imagine a new class of engineered sensing systems that have general processing units along the lines of the learning mechanism in this paper, that enables perceptual inference from a broader range of sensory streams that do not require the precision manufacturing common to many current sensors and cameras. 

The learning notion of perception emerging from a process of disentangling it from encoded sensory streams is an interesting intellectual view of how perception may generally emerge in the brain. 
%Abstracting the ideas in the current paper, we can see a generalizable mathematical structure. 
Any sensory stream going to the brain encodes  information from the world entangled with sensing organ characteristics. The self-supervised learning mechanism proposed in this paper might be abstracted into a general neural process of learning to predict fluctuations in sensory stream due to ego perturbation, 
%within a decode, percept-perturbations, re-encode pipeline as in this paper. And a neural search for a perceptual representation with associated pipeline that results in accurate prediction, with optima defining what a percept ``is'' for that sensory stream. 
with perception for that sense emerging neurally as the optimal internal representation and associated decoder/encoder pair that enables accurate prediction. 

To conclude, we invite the research community to build on the computational framework proposed in this paper, by improving the visual system components (e.g. even more accurately modeling details in the eye model, or replacing back-propagation in the cortical model with more bio-plausible learning rules~\citep{lillicrap2020backpropagation, hinton2022forward}), or applying the framework to other sensory modalities. 

%% file: 08-Acknowledgement.tex
\section*{Acknowledgements}
This project was supported by Air Force Office of Scientific Research's MURI grant (FA9550-20-1-0195 and FA9550-21-1-0230). 
We are grateful to everyone involved in the MURI project for their support and feedback. 
Special thanks to Bruno Olshausen, Jessica Lee, and Alexander Belsten for their valuable discussions and insights.  
We also appreciate Fred Rieke, Lawrence Sincich, and David Brainard for their thoughtful input and engaging conversations.

%% file: S-01-intro.tex
In this supplementary section, we provide additional details on our computational framework for modeling the emergence of human color vision. Section~\ref{sec:supp-retina} covers the simulation of optic nerve signals from hyperspectral scene images and our modeling of retina circuitry.  Section~\ref{sec:supp-cortex} details the design of our simulated cortical model.  Section~\ref{sec:supp-cmfsim} provides a step-by-step description of how we simulate formal color matching function tests of color dimensionality, \CMFSim. In Section~\ref{sec:supp-baseline}, we probe whether the emergence of color vision depends on cortical learning, by introducing comparative, baseline cortical models.
Section~\ref{sec:supp-boosting} simulates variant cell expression phenotypes after gene therapy in squirrel monkeys~\citep{mancuso2009gene}, showing robustness of the simulation result that addition of a new cell opsin gene enhances color dimensionality. Finally, Section~\ref{sec:supp-tetrachromacy} discusses our tetrachromacy simulation, highlighting the differences between the human Q cone (between M and L cones) and the pigeon Q cone (between S and M cones).
% Upon acceptance, we will make our entire codebase available along with trained model weights and comprehensive documentation.
A supplementary video is available on our project website: \url{https://matisse.eecs.berkeley.edu}.

\begin{table}[h]
    \newcommand{\glossarytablewidth}{1.0\columnwidth}
    \newcommand{\leftcolumnwidth}{1.7cm}
    \newcommand{\subtablegap}{0.3em}
    \newcolumntype{C}{>{\centering\arraybackslash}p{\leftcolumnwidth}}
    
    \begin{center}
    \caption{Glossary of terms used in our framework}\label{table:glossary}%
    % \vspace{1em}
    
    \begin{tabularx}{\glossarytablewidth}{|C|X|}
    \hline
    \multicolumn{2}{|c|}{Scalars} \\
    \hline
    $t, x, y$   & Time, world view coordinates    \\
    $u,v$       & Optic nerve image coordinates  \\
    $\ColorPerceptDim$         & $\ColorPerceptDim$-dimensional vector represents cortical color\\ 
    \hline
    \end{tabularx}
    
    \vspace{\subtablegap}

    \begin{tabularx}{\glossarytablewidth}{|C|X|}
    \hline
    \multicolumn{2}{|c|}{Eye Simulation Engine} \\
    \hline
    $\SceneImage{t}(x,y)$   & Scene image, \SPD at each pixel    \\
    $\RetinaEncodingOperator$   & Retina encoding of scene into time-averaged optic nerve image: $\RetinaEncodingOperator(\SceneImage{t}) = \OpticNerveSignal{t}$\\ 
    \hline
    \end{tabularx}
    
    \vspace{\subtablegap}

    \begin{tabularx}{\glossarytablewidth}{|C|X|}
    \hline
    \multicolumn{2}{|c|}{Cortical Image Functions} \\
    \hline
    $\OpticNerveSignal{t}(u,v)$        & Time-averaged optic nerve image\\ 
    %$\ONSBlurred{t}(u,v)$        & Blurred optic nerve image signal                        \\
    $\PhotoreceptorActivations{t}(u,v)$        & Cone activations image\\ 
    $\VisualPerceptDistorted{t}(u,v)$        & Visual percept image \\
    $\VisualPercept{t}(x,y)$        & Unwarped visual percept image\\
    $\PredictedOpticNerveSignal{t+dt}(u,v)$       & Predicted optic nerve image, time $t+dt$\\
    \hline
    \end{tabularx}
    
    \vspace{\subtablegap}
    
    \begin{tabularx}{\glossarytablewidth}{|C|X|}
    \hline
    \multicolumn{2}{|c|}{Neural Buckets (Learnable Parameters)} \\
    \hline
    $\ConeTypesBucket$        & Color type of each cone                        \\
    $\ConePositionsBucket$        & Position of each cone in visual field            \\
    $\LateralInhibitionWeightsBucket$        & Lateral inhibition weights to neighboring cones       \\
    $\DemosaickerBucket$        & Demosaicing operator parameters       \\
    $\MotionEstimationBucket$        & Motion estimation operator parameters       \\
    \hline
    \end{tabularx}
    
    \vspace{\subtablegap}
    
    \begin{tabularx}{\glossarytablewidth}{|C|X|}
    \hline
    \multicolumn{2}{|c|}{Learned Cortical Function Operators} \\
    \hline
    $\DecodeOperator$, $\EncodeOperator$  & $\DecodeOperator$ decodes $\OpticNerveSignal{t}$ into $\VisualPercept{t}$, and $\EncodeOperator$ re-encodes $\VisualPercept{t}$ into $\OpticNerveSignal{t}$. \\
    & $\DecodeOperator = \DecodeOperator_{\ConePositionsBucket} \circ \DecodeOperator_{\ConeTypesBucket} \circ \DecodeOperator_{\LateralInhibitionWeightsBucket}$, and $\EncodeOperator = \EncodeOperator_{\LateralInhibitionWeightsBucket} \circ \EncodeOperator_{\ConeTypesBucket} \circ \EncodeOperator_{\ConePositionsBucket}$. \\
    $\DecodeOperator_{\LateralInhibitionWeightsBucket}$, $\EncodeOperator_{\LateralInhibitionWeightsBucket}$  & Transforms $\OpticNerveSignal{t}$ into $\PhotoreceptorActivations{t}$, and vice-versa.\\
    $\DecodeOperator_{\ConeTypesBucket}$, $\EncodeOperator_{\ConeTypesBucket}$ & Transforms $\PhotoreceptorActivations{t}$ into $\VisualPerceptDistorted{t}$, and vice versa.   \\
    $\DecodeOperator_{\ConePositionsBucket}$, $\EncodeOperator_{\ConePositionsBucket}$ & Transforms $\VisualPerceptDistorted{t}$ into $\VisualPercept{t}$, and vice versa.   \\
    \hline
    $\MotionEstimationFunctionTemplate{\OpticNerveSignal{1}}{\OpticNerveSignal{2}}$ & Estimates world translation $(x,y)$ between two optic nerve images \\
    $\TranslateOperator_{(x,y)}$ & Translates an image laterally by $(x,y)$. $\TranslateOperator_{\MotionEstimationFunction}(\VisualPercept{t})=\VisualPercept{t+dt}$   \\
    \hline
    \end{tabularx} 
    \end{center}
\end{table}

%% file: S-02-retina-model.tex
\section{Simulation Engine of Biological Eyes}\label{sec:supp-retina}

% \inlineren{Section intro paragraph. Reference figure images and video timestamps.}

We simulate the optic nerve signals from hyperspectral scene images, simulate eye gaze movements, and model the retina circuitry. We describe the details of these simulations in the following.

\subsection{Database of Hyperspectral Scene Images}
\label{sec:methods-scene-images}

% We model the scene viewed by the eye using image databases comprising $900$ photographs of everyday scenes shot with a hyperspectral camera~\citep{arad2022ntire} where each pixel is represented by spectral power distributions (``\SPDs'' below). 
% . This avoids assumptions about color dimensionality, allowing cortical learning to infer dimensionality freely and making the task more challenging and unbiased 
We model the scene viewed by the eye using a dataset comprising 900 hyperspectral images from everyday scenes, captured with a hyperspectral camera~\citep{arad2022ntire}. Each image in the dataset is of dimension 512 pixels × 482 pixels × 31 spectral channels, where each pixel is represented by its spectral power distribution (``\SPDs’’ below). To generate our training data set, we cropped thousands of 482x482 patches from these hyperspectral images, ensuring sufficient resolution to avoid aliasing effects, induced by varying cone cell densities (Section~\ref{sec:methods-foveation-cone-positions}).

While only hyperspectral images are used during learning simulation, RGB images can be presented during test-time by simulating a projector with specific spectral power distributions (SPDs) for the R, G, and B channels~\citep{cottaris2019computational}, effectively converting each RGB image into a spectral representation before passing into the retina model. This test-time use of RGB images does not affect learning and is purely for illustrative purposes in the figures (e.g., Figure~\ref{fig:3}).

% \TODO{Discuss monochromatic color patches.}

\subsection{Eye Gaze Movements -- Fixational Drift}
We simulate small eye movement as fixational eye drift with a random walk~\citep{young2021emulated}. This produces a stream of the current scene image translating across the retina.  Over a time interval of duration $dt$, we sample the change in eye motion from a uniform probability distribution $(dX, dY) \sim \mathcal{U}\{(-15,15)\times (-15, 15)\}$ where $15$ corresponds to $15$ pixels in the scene image $S$. 

The Mona Lisa video in Supplementary Video 2:08 also incorporates manually-authored saccades (e.g.\ from hand to mouth) for illustrative purposes, but the learning simulation uses only fixational drift.

\subsection{Retina Circuitry Model}

Our model of light detection and signal encoding in the retina is the textbook model~\citep{rodieck1998first} of the photopic, midget, ``private-line'' pathway (see Figure~\ref{fig:2}). Details of our implementation follow. 

\subsubsection{Cone Mosaic and Spectral Sampling}
\label{sec:methods-cone-mosaic}
In our simulation, we model $256\times 256$ cone cells, with the random distribution of L, M and S cone cells on the retina with a relative probability of $0.63$, $0.32$ and $0.05$ from \citet{sabesan2015characterizing}. We model L, M and S spectral response using the template response function from \citet{carroll2000flicker}, with spectral peaks of $560$, $530$, $419$ nm, respectively. 
% For the tetrachromatic retina, we model 50\% of L cells as a fourth ``Q'' cone, using a spectral response function given by Carroll~et~al.~\cite{carroll2002estimates} with a spectral peak of $514$ nm. 
We model photon-shot noise in the photoreceptor activation values, simulating a signal-to-noise ratio of approximately $100$.  The simulation omits foveal tritanopia~\citep{williams1981foveal}.

\subsubsection{Foveation and Cone Positions}\label{sec:methods-foveation-cone-positions}
We model cone cell spatial density on the retina in accordance with \citet{curcio1990human}, while enforcing retinotopy.  Starting from a regular grid of cells, we computationally perturb positions until the target spatial distribution is stochastically achieved, while constraining each cell to be always confined by its neighbors. We model biological variation in cell locations, which represents a challenge for cortical learning, by adding multi-resolution positional randomness~\citep{perlin1985image}. 

\subsubsection{Center-Surround Lateral Inhibition}
\label{sec:methods-retina-LI}

We mathematically model textbook lateral inhibition in retinal signals as a Difference-of-Gaussians (DoG) convolution kernel~\citep{enroth1983spatio}, with parameters fitted from electrophysiology~\citep{wool2018nonselective}. Specifically our DoG kernel has standard deviations of $0.15$ and $0.9$ cone diameters, respectively, for the positive center and negative surround Gaussians. Significantly, the relative amplitudes are such that the kernel has a non-zero mean of $0.09$~\citep{lennie1991design,wool2018nonselective}. 
We convolve the array of cone activation values by this DoG kernel to compute outputs to bipolar cells.

\subsubsection{On/Off Pathways and Optic Nerve Spiking}
\label{sec:methods-retina-on-off}
We model textbook on- and off- connections from cone activations to bipolar cells, forwarding to on- and off- retinal ganglion cells (RGC)s. The on- activation is modeled by a rectified linear unit activation function, $\mathrm{ReLU}(x)=\mathrm{max}(0,x)$, where $x$ is the laterally-inhibited cone output. The off- activation is modeled as $\mathrm{ReLU}(-x)$. 
Optic nerve spikes are the outputs at RGC axons, with action potentials are generated with Leaky Integrate and Fire model~\citep{lapicque1907recherches,abbott1999lapicque,arbib2003handbook}. 

%% file: S-03-cortical-model.tex
\section{Cortical Model}\label{sec:supp-cortex}

\subsection{Time-Averaged Optic Nerve Signals}
\label{sec:methods-time-averaging}

We model the first step of cortical processing as re-combination of the on- and off- RGC action potentials followed by time-averaging.  This results in an image stream $\OpticNerveSignal{t}$ with real-valued pixel values that are proportional to the laterally-inhibited output from cone cells.  $\OpticNerveSignal{t}$ is the sole input to the remainder of the cortical model.

%\subsection{Prediction Pipeline Functions \& Neural Buckets}
\subsection{Prediction Pipeline and Learning Objective}
\label{sec:methods-pipeline}

Figure~\ref{fig:7} shows an elaboration of the cortical prediction pipeline introduced in Figure~\ref{fig:3}.1 and summarized in Section~\ref{sec:method_learning}. This pipeline represents an existence-proof simulation that the hypothesized self-supervised learning on $\OpticNerveSignal{t}$ can successfully produce: (1) emergence of vision with the correct color dimensionality, and (2) inference of invariant retinal properties, including learned neural buckets for cone spectral identities $\ConeTypesBucket$, cell positions $\ConePositionsBucket$ and lateral inhibition weights $\LateralInhibitionWeightsBucket$. 
Our results show that both can be simultaneously achieved by learning to minimize the error in predicting the fluctuations in optic nerve signal values. 

To re-summarize Section~\ref{sec:method_learning}, the prediction is made by applying the learned functions for decoding $\DecodeOperator$, translation $\TranslateOperator$ and re-encoding $\EncodeOperator$ to the optic nerve signal $\OpticNerveSignal{t}$, to obtain a predicted optic nerve signal $\PredictedOpticNerveSignal{t+dt} = \EncodeOperator(\TranslateOperator( \DecodeOperator(\OpticNerveSignal{t})))$.  The learning objective function is to minimize prediction error, the difference between predicted and real optic nerve images at time $t+dt$: $E_\mathrm{prediction} = \| \OpticNerveSignal{t+dt} - \PredictedOpticNerveSignal{t+dt} \|_2^2 
    = \| \OpticNerveSignal{t+dt} - \EncodeOperator(\TranslateOperator( \DecodeOperator(\OpticNerveSignal{t}))) \|_2^2.$
The decoder $\DecodeOperator$ and re-encoder $\EncodeOperator$ functions are each factorized into a pipeline, such that: 
$\DecodeOperator = \DecodeOperator_\ConePositionsBucket \circ \DecodeOperator_\ConeTypesBucket \circ \DecodeOperator_\LateralInhibitionWeightsBucket$ and $\EncodeOperator = \EncodeOperator_\LateralInhibitionWeightsBucket \circ \EncodeOperator_\ConeTypesBucket \circ \EncodeOperator_\ConePositionsBucket$, where each sub-function is an operator conditioned on its corresponding neural bucket, $\ConeTypesBucket$, $\ConePositionsBucket$ and $\LateralInhibitionWeightsBucket$.  The following sections describe the implementation of each learnable neural bucket and its associated cortical sub-function. 
% 
% \begin{align}
%     E_\mathrm{prediction} &= \| \OpticNerveSignal{t+dt} - \PredictedOpticNerveSignal{t+dt} \|_2^2 \\
%     &= \| \OpticNerveSignal{t+dt} - \EncodeOperator(\TranslateOperator( \DecodeOperator(\OpticNerveSignal{t}))) \|_2^2. \label{eq:prediction_error}
% \end{align}

\begin{figure}[t!]
   \includegraphics[width=\columnwidth]{assets/fig7_0522.pdf}
    \caption{Pipeline of decoding, translation and re-encoding cortical functions. The optic nerve signal image $\OpticNerveSignal{t}$ is decoded into visual percept image $\VisualPercept{t}$. The percept is translated in accordance with eye motion over time $dt$, then re-encoded back into a prediction of the optic nerve signal $\PredictedOpticNerveSignal{t+dt}$ at a short time in the future.}
    \label{fig:7}
\end{figure}

\subsubsection{Learning Lateral Inhibition Neighbor Weights $\LateralInhibitionWeightsBucket$}
\label{sec:methods-cortex-LI}

In the factorized function pipeline, sub-decoder $\DecodeOperator_\LateralInhibitionWeightsBucket$ conceptually inverts lateral inhibition to transform optic nerve image $\OpticNerveSignal{t}$ into cone activation image $\PhotoreceptorActivations{t}$. Conversely, sub-encoder $\EncodeOperator_\LateralInhibitionWeightsBucket$ reproduces lateral inhibition to transform $\PhotoreceptorActivations{t}$ into $\OpticNerveSignal{t}$.  These operators are modeled, respectively, as inverse and forward convolution operators, implemented by pixel-wise division and multiplication in the Fourier domain. The learnable parameters for this operator are neural bucket $\LateralInhibitionWeightsBucket$, representing the Fourier transform image pixels of the lateral inhibition DoG kernel's Fourier transform. The resolution of $\LateralInhibitionWeightsBucket$ is set equal to the image resolution of $\OpticNerveSignal{t}$.

\subsubsection{Learning Color Types $\ConeTypesBucket$ and Interpolation $\DemosaickerBucket$}
\label{sec:methods-cortex-C}

In the factorized function pipeline, sub-decoder $\DecodeOperator_\ConeTypesBucket$ conceptually transforms cone activation image $\PhotoreceptorActivations{t}$ (scalar-valued pixels) into a full-color visual percept image $\VisualPerceptDistorted{t}$ ($\ColorPerceptDim$-dimensional vector pixels). Conversely, $\EncodeOperator_\ConeTypesBucket$ re-encodes $\VisualPerceptDistorted{t}$ into $\PhotoreceptorActivations{t}$.  These functions depend on two neural buckets of learnable parameters, $\ConeTypesBucket$ and $\DemosaickerBucket$. $\ConeTypesBucket$ is an image of $\ColorPerceptDim$-dimensional vectors, in which the cortical model learns to represent the spectral type of the source cone associated with each pixel in $\PhotoreceptorActivations{t}$. $\DemosaickerBucket$ are the parameter weights of a learnable function $\phi_\DemosaickerBucket$ that spatially interpolates color across the image, which we implement as a convolutional neural network based on the U-Net~\citep{ronneberger2015u} architecture. 
% This gives $\DecodeOperator_\ConeTypesBucket$ and $\EncodeOperator_\ConeTypesBucket$ defined as:
% \begin{align}
%     \DecodeOperator_\ConeTypesBucket(\PhotoreceptorActivations{t}) &= \phi_\DemosaickerBucket(\PhotoreceptorActivations{t} \otimes \ConeTypesBucket),\\
%     \mathrm{and\ } \EncodeOperator_\ConeTypesBucket(\VisualPerceptDistorted{t}) &= \VisualPerceptDistorted{t} \otimes \ConeTypesBucket),
% \end{align}

Mathematically, $\DecodeOperator_\ConeTypesBucket$ and $\EncodeOperator_\ConeTypesBucket$ are defined as $\DecodeOperator_\ConeTypesBucket(\PhotoreceptorActivations{t}) = \phi_\DemosaickerBucket(\PhotoreceptorActivations{t} \otimes \ConeTypesBucket)$,
where $\otimes$ denotes element-wise multiplication; $\EncodeOperator_\ConeTypesBucket(\VisualPerceptDistorted{t}) = \VisualPerceptDistorted{t} \otimes \ConeTypesBucket$, where $\odot$ denotes element-wise dot product. To maintain $\DecodeOperator_\ConeTypesBucket$ and $\EncodeOperator_\ConeTypesBucket$ as pseudo-inverses, we constrain $\ConeTypesBucket \odot \ConeTypesBucket$ to be an image with all pixels equal to 1. 

\subsubsection{Learning Cell Positions $\ConePositionsBucket$}
\label{sec:methods-cortex-P}

As shown in Fig.~\ref{fig:2}.3.F, optic nerve images are spatially distorted due to foveation, whereas visual perception is apparently undistorted.  
%These include image signals $\OpticNerveSignal{t}$, $\PhotoreceptorActivations{t}$ and full-color visual percept $\VisualPerceptDistorted{t}$.  
In the factorized function pipeline, sub-decoder $\DecodeOperator_\ConePositionsBucket$ conceptually transforms $\VisualPerceptDistorted{t}(u,v)$, which is foveated and spatially distorted, into the final visual percept image $\VisualPercept{t}(x,y)$, which is Euclidean and undistorted.  Conversely, $\EncodeOperator_\ConePositionsBucket$ re-warps  $\VisualPercept{t}(x,y)$ into $\VisualPerceptDistorted{t}(u,v)$.

We define a learnable spatial warping function $\phi_\ConePositionsBucket(u,v)=(x,y)$, implemented using normalizing flow~\citep{rezende2015variational,dinh2016density} that is an invertible function.  Neural bucket $\ConePositionsBucket$ comprises the learnable parameters of this normalizing flow network. Then $\DecodeOperator_\ConePositionsBucket(\VisualPerceptDistorted{t})(x,y)=\VisualPerceptDistorted{t}(\phi_\ConePositionsBucket^{-1}(x,y))$, and $\EncodeOperator_\ConePositionsBucket(\VisualPercept{t})(u,v)=\VisualPerceptDistorted{t}(\phi_\ConePositionsBucket(u,v))$.

\subsection{Learning Eye Motion Estimation $\MotionEstimationBucket$}
\label{sec:methods-cortex-M}

In the learning loop, cortical sub-function $\TranslateOperator$ translates the visual percept image according to the spatial motion $(dx,dy)$ that occurs during the brief period between $t$ and $t+dt$. Where does $(dx,dy)$ come from? Here, we describe how the cortex can learn a helper cortical function $\MotionEstimationFunction$ that estimates $(dx,dy)$ directly from the optic nerve stream values at times $t$ and $t+dt$, maintaining the strict operation of the cortical learning model purely from optic nerve signals. 
% This pure version is presented in Fig.~\ref{fig:3} experiments, showing that overall color vision, including motion estimation, can be learned as hypothesized.  For computational efficiency in most experiments, we provide an oracle function $\MotionOracleFunction$ that gives the ground truth $(dx,dy)$ translation from eye gaze movements used in the retina model encoding. 

We model learning of $\MotionEstimationFunction$ with neural bucket $\MotionEstimationBucket$ by optimizing:
\newcommand{\Blur}{\mathrm{B}}
\begin{align}
    \MotionEstimationBucket = \underset{\MotionEstimationBucket,\ConePositionsBucket}{\mathrm{argmin}} \| \Blur(\OpticNerveSignal{t+dt}) -  \EncodeOperator_\ConePositionsBucket(\TranslateOperator_{\MotionEstimationFunction}( \DecodeOperator_\ConePositionsBucket(\Blur(\OpticNerveSignal{t}))))\|_2^2, \nonumber
\end{align}
where $\Blur$ is a convolution operator that low-pass filters the optic nerve images, effectively halving the highest frequencies; $\DecodeOperator_\ConePositionsBucket$ aims to dewarp the optic nerve image into Euclidean coordinates, and $\EncodeOperator_\ConePositionsBucket$ is the inverse operator that rewarps the image. Details are discussed also in Supplementary Video 16:25.

% ; and In sandbox experiments, where the retina model does not contain foveation, eye motion estimation 

\subsection{Efficient Learning Implementation \& Alternative Representation of Internal Percepts}
\label{sec:methods-learning-implementation}

We simulate the self-supervised learning by parallel numerical optimization of all neural buckets
by applying the stochastic gradient descent algorithm to minimize the prediction error metric. In practice, this is implemented by repeating the following step thousands of times until convergence: pick a batch of different scene images from the database; generate the optic nerve signals at time $t$ with the retina encoding model; send this into the cortical model and execute the decode / translate / re-encode functions with current neural bucket parameters to estimate the signal at $t+dt$; compute the actual optic nerve signal at $t+dt$ with the retina encoding model; compute the prediction error by taking the difference between prediction and actual signal; backpropagate the prediction error to update neural bucket parameters. 

For efficient processing of cortical functions, we avoid computing full-resolution undistorted visual percept images $\VisualPercept{t}(x,y)$ during the learning simulation.  Instead, we directly compute $\VisualPerceptDistorted{t+dt}(u,v)$ by optical flow of the pixels in $\VisualPerceptDistorted{t}(u,v)$, where the optical flow map is defined by $(\EncodeOperator_\ConePositionsBucket \circ \TranslateOperator_{\MotionEstimationFunction} \circ \DecodeOperator_\ConePositionsBucket)$.  This is mathematically equivalent to the learning implementation described in Section~\ref{sec:methods-pipeline}, 
% but leads to $N$x improvement in memory consumption and $K$x in training duration.\footnote{\ak{what are $N$ and $K$ here?}} 
and the cortical model continues to learn cell positions $\ConePositionsBucket$, and therefore decoder $\DecodeOperator$ can still be applied to compute undistorted visual percepts when desired.

% \subsection{Alternative Model Architecture}\label{sec:cortex-alternative}

%% file: S-04-CMF-SIM.tex
\begin{algorithm}[t!]\label{algo:cmf-sim}
\caption{Pseudocode for \CMFSim}

\DontPrintSemicolon

\KwData{WAVELENGTHS, MAX\_TRIALS, Retinal process $\Upsilon$, Cortical decoder $\Phi$}

\;

\SetKwFunction{FCMFSIM}{CMF-SIM}
\SetKwFunction{FComputeBaseErrors}{ComputeBaseErrors}
\SetKwFunction{FFindMinimumPrimaries}{FindMinimumPrimaries}
\SetKwProg{Fn}{Function}{:}{}

\Fn{\FCMFSIM{}}{
    base\_errors $\gets$ \FComputeBaseErrors{}\;
    color\_dimensionality $\gets$ \FFindMinimumPrimaries{base\_errors}\;
    \KwRet color\_dimensionality\;
}

\; 
\Fn{\FComputeBaseErrors{}}{
    Initialize an array errors\;
    \ForEach{wavelength $\lambda$ in WAVELENGTHS}{
        $(x,y)_A$ = RandomLocation()\;
        
        $(x,y)_B$ = RandomLocation()\;
        
        colorPatchA $\gets$ MonochromaticColorPatch($\lambda$, $(x,y)_A$ )\;
        
        colorPatchB $\gets$ MonochromaticColorPatch($\lambda$, $(x,y)_B$ )\;
        
        perceptA $\gets \Phi(\Upsilon(\text{colorPatchA}))$\;
        
        perceptB $\gets \Phi(\Upsilon(\text{colorPatchB}))$\;
        errors[$\lambda$] $\gets$ PerceptualError(perceptA, $(x,y)_A$, perceptB, $(x,y)_B$ )\;
    }
    \KwRet errors\;
}

\;

\Fn{\FFindMinimumPrimaries{base\_errors}}{
    num\_primaries $\gets 0$\;
    \Repeat{all\_tests\_passed}{
        num\_primaries $\gets$ num\_primaries + 1\;
        all\_tests\_passed $\gets$ true\;
        trial $\gets$ 0\;
        \Repeat{trial $=$ MAX\_TRIALS}{
            trial $\gets$ trial + 1\;
            Randomly initialize primaries $p=\{p_1, ..., p_{\text{num\_primaries}}\}$\;
            \ForEach{wavelength $\lambda$ in WAVELENGTHS}{
                Zero initialize coefficients $\alpha = \{\alpha_1, ..., \alpha_{\text{num\_primaries}}\}$ for primary $p$\;
                \Repeat{convergence}{
                    $(x,y)_A$, $(x,y)_B$  = RandomLocations()\;

                    positive\_alpha, negative\_alpha $\gets$ $\alpha_i^+$ for all $i$, $(-\alpha_i)^+$ for all $i$\;
                    
                    testColorPatch $\gets$ MonochromaticColorPatch($\lambda$, $(x,y)_A$ + WeightedColorPatch(negative\_alpha, $p$, $(x,y)_A$)\;
                    
                    matchColorPatch $\gets$ WeightedColorPatch(positive\_alpha, $p$, $(x,y)_B$)\;
                    testPercept $\gets \Phi(\Upsilon(\text{testColorPatch}))$\;
        
                    matchPercept $\gets \Phi(\Upsilon(\text{matchColorPatch}))$\;
                    errors[$\lambda$] $\gets$ PerceptualError(testPercept, $(x,y)_A$, matchPercept, $(x,y)_B$ )\;
                    coefficients $\alpha$ $\gets$ GradientDescent($p$, $\alpha$, target, match, errors)\;
                }
                \If{error $\geq$ base\_errors[$\lambda$]}{
                    all\_tests\_passed $\gets$ false\;
                    break\;
                }
            }
        }
    }
    \KwRet num\_primaries\;
}

\end{algorithm}

\section{Color Matching Function Tests \& \CMFSim}\label{sec:supp-cmfsim}

In the classical color matching theory of~\cite{maxwell1856theory} and~\cite{grassmann1853theorie}, an observer compares a test color \SPD, $t$, and a set of $K$ primary color \SPDs, $p_i$ for $i=1,..., K$ and tunes real weights $\alpha_i$ for each primary until a color match is achieved. If a particular weight has a negative value, primary light of that magnitude is added to the test color $t$ rather than to the other primary colors.  Mathematically, a color match is achieved when the matching \SPD $\sum_{i=1}^K\mathrm\alpha_i^+ p_i$ and test \SPD $t+\sum_{i=1}^K(-\alpha_i)^+p_i$ appear identical. Here, $\alpha_i^+$ is defined as $\max(\alpha_i, 0)$.

We simulate such classical color matching by formulating the retina model plus learned cortical model as a black-box color observer in the colorimetric sense. That is, we take the test and matching \SPDs, simulate the stimulation on different parts of the retina for test and match color patches, apply the retina encoding model $\RetinaEncodingOperator$ and the learned cortical decoder $\DecodeOperator$ to each separately, and iteratively update the weights $\alpha_i$ until there is no further improvement in the computed perceptual difference $E=\|\DecodeOperator(\RetinaEncodingOperator(\sum_{i=1}^K\mathrm\alpha_i^+ p_i))-\DecodeOperator(\RetinaEncodingOperator(t+\sum_{i=1}^K(-\alpha_i)^+p_i))\|_2^2$. Note that the optimal $E$ will not be close to zero if the primaries cannot match test color $t$.

The color dimensionality of an observer is formally equal to the minimum number of primary colors required to match any test color. \CMFSim simulates thousands of color matches to rigorously compute this dimensionality for any given retina model plus cortical model.  In \CMFSim, we simulate classical color matching function (CMF) tests of exhaustively attempting to match test colors equal to each monochromatic wavelength (100 samples from 400 to 700 nm), with a linear combination of $K$ primary colors. For the primary colors, we use distinct, monochromatic \SPDs.  We start CMF tests with a single primary, and add primaries one-by-one until all test wavelengths in the CMF can be matched successfully. The pseudocode for \CMFSim is described in Algorithm~\ref{algo:cmf-sim}.

For example, for our cortical model to be formally measured as trichromat, we must show that there do not exist any 1 or 2 primaries that can pass the CMF tests, and show 3 primaries that succeed. A brute-force approach would be to exhaustively test all possible sets of $K$ primaries; for efficiency, we instead perform stratified sampling to randomly generate a large number (e.g.~500) distinct sets of primaries for each choice of $K$. We require that all of these possible primary sets fail, with the aggregate perceptual error across all wavelengths being greater than a threshold $\epsilon$, before incrementing the CMF tests to $N$+1 primaries.

%% file: S-05-baseline.tex
\begin{figure}[t!]
    \centering
    \includegraphics[width=0.7\columnwidth]{assets/fig8_0520.pdf}
    \caption{Comparison \CMFSim results of our baseline cortical models with limited or no inferential processing.  In all of these models, the input optic nerve stream comes from a retina with 3 cone cells as in regular human trichromacy.  1. The first baseline model represents  no cortical learning involvement in human visual perception, by setting the internal percept directly equal to the input optic nerve signal. The resulting base error, the perceptual error between the same color patch at two different retinal locations, results in higher errors, compared to the energy of the internal percept. This means that this cortical model fails to recognize the same color consistently across different patches of the retina, so it is procedurally impossible to even try color matching function tests against reference match colors. Formally then, this model fails resolve color vision. 2. In the second baseline model, we treat cone cell activations as internal percepts, omitting lateral inhibition from the eye simulation; \CMFSim measures the resulting internal percept as 1D color vision, failing to resolve trichromacy. 3. The third baseline model is an ablation model in which we omit the demosaicking network $\DemosaickerBucket$ from our proposed model of Section~\ref{sec:supp-cortex}; \CMFSim also measures the resulting internal percept as 1D color. Red base errors are superimposed on energy / 1 primary error plots for visual comparison, and 150\% of these base errors are applied as threshold determination.}
    \label{fig:8}
\end{figure}

\section{Baseline Results -- Testing if Cortical Inference is Required for Color Vision}\label{sec:supp-baseline}

A fundamental question about color perception is whether cortical inference is necessary, or if human color perception arises directly from hardwired neural circuits. The main paper presents analysis an existence proof that cortical inference can indeed result in color vision of the correct dimensionality.  Here, we add evidence that cortical inference of some kind is necessary, by modeling and testing three baseline cortical models with limited or no cortical processing.

For our first baseline, we assume an extreme case where no cortical learning occurs, treating raw optic nerve signals  directly as internal percepts (i.e., the cortical decoder function $\Phi$ is an identity function). We apply \CMFSim directly to the optic nerve signals generated by a trichromat retina. Figure~\ref{fig:8}.1 presents analysis that this model fails to generate vision that can be recognized as color consistent. The analysis is to perform baseline color matching experiments where the test and match spectral functions are identical.  This ``base perceptual error'' is larger than the energy of each color percept itself, meaning that the emergent vision fails to recognize a patch as the same color across different parts of the retina. 

The second baseline model is a variant of the first, where we change the optic nerve stream to directly transfer cone activation values, omitting the encoding complications of foveated warping and lateral inhibition. This is intended as a far simpler encoding, to pressure test whether color vision can be detected in the spectrally encoded cone values without further processing. However, \CMFSim formally measures that this model results in 1D color (Figure~\ref{fig:8}.2), instead of the expected 3D color from such a retina. This result shows that cortical processing is required even on cone activations to produce color vision of the correct color dimensionality.

The third baseline model adds another perspective by taking the full-featured model detailed in Section~\ref{sec:supp-cortex} but omitting the demosaicking network $\phi_\DemosaickerBucket$ (as defined in Section~\ref{sec:methods-cortex-C}).  As shown in Figure~\ref{fig:8}.3, \CMFSim measures the resulting percepts of this model also as 1D, providing evidence of the importance of demosaicking process in the emergence of color vision.

%% file: S-06-gene-therapy.tex
\section{Variations in Cell Expression after Gene Therapy}\label{sec:supp-boosting}

%In gene therapy experiments of boosting color dimensionality~\cite{mancuso2009gene}, additional cone types are added to the retina in adulthood, and boosting of color dimension is 
As described in Section~\ref{sec:result_emergence}, we simulate experiments aimed at boosting color dimensionality~\cite{mancuso2009gene}. In our simulation, approximately 60\% of M cones are affected by gene therapy, with three possible scenarios for their modification. First, affected M cones are completely transformed into pure L cones (Figure~\ref{fig:9}.1). Second, affected M cones equally express M and L opsins (Figure~\ref{fig:9}.2). Third, affected M cones express M and L opsins in random ratios, that is $\alpha$L $+$ $\beta$M, where $\alpha$ and $\beta$ spatially vary across the retina (Figure~\ref{fig:9}.3). In all cases, we confirm that our cortical model acquires 3D color vision after adaptation (i.e. re-learning), formally measured by \CMFSim.
This makes intuitive sense if we consider the emergent vision from linear systems theory. Each different opsin can be interpreted as a basis vector for the resulting linear color space. In this view, cells containing a mixture of two cones (M and L in this case) can be interpreted as linear combinations of two basis vectors, which cannot increase the dimensionality of the linear space. 

\begin{figure}[t!]
    \centering
    \includegraphics[width=\columnwidth]{assets/fig9_0827.pdf}
    \caption{Three scenarios of cell expression after gene therapy adding a third cone type: 1. affected M cones transform into L cones, 2. affected M cones express 50:50 M and L opsins, and 3. affected M cones express M and L opsins at spatially-varying, random ratios (i.e. $\alpha, \beta \in [0,1], \alpha+\beta = 1$). \CMFSim shows that all cases converge to 3D color vision after post-therapy re-learning.}
    \label{fig:9}
\end{figure}

%% file: S-07-tetrachromacy.tex
\section{Probing Effect of Spectral Response and Environmental Colors in Acquiring Tetrachromatic Vision}\label{sec:supp-tetrachromacy}

In Section~\ref{sec:result_emergence}, tests of tetrachromatic eyes assumed the fourth cone type was a pigeon Q cone, with peak sensitivity at 506 nm, between the S and M cone peak sensitivities (419 nm and 530 nm, respectively). In this section, we report additional experiments using instead a human Q cone, modeled with peak sensitivity at 545 nm between M and L peak sensitivities (530 nm and 560 nm, respectively).  As shown in the spectral graphs of Figure~\ref{fig:10}, the pigeon Q cone is more decorrelated from L, M, S, which in principle may more easily support emergence of 4D color vision.

Further, we probe the effect of visual environment on the emergence of tetrachromacy, simulating the learning of color vision with three different input scene image datasets. The first environment is the set of real hyperspectral photographs of everyday scenes~\citep{arad2022ntire} as described in Section~\ref{sec:methods-scene-images}. However, natural hyperspectral images are thought to be lacking in human tetrachromatic colors~\citep{leeTetrachromacy2024}, so in the second and third datasets we augment the 900 hyperspectral images with 100 tetrachromatic color patches. In the second dataset, the colors are sampled from the tetrachromatic hue sphere as computed using recently developed $N$-dimensional color theory for tetrachromacy~\citep{leeTetrachromacy2024}, customized to each of the observers here (i.e. with human Q or pigeon Q, respectively).  In the third dataset, the colors are sampled from an idealized tetrachromatic color space in which each of the four cone channels is allowed to take any value, allowing theoretically maximum levels of chromatic contrast.  Since natural hyperspectral images have been found lacking in human tetrachromatic colors~\citep{leeTetrachromacy2024}, in principle we may expect the likelihood that 4D color vision emerges to increase across these three simulated environments.

%The second and third are synthetic datasets, composed of randomly overlapping, colored polygons\TODO{~\cite{deadleaves}}. In the second dataset, the colors of the polygons are sampled from the tetrachromatic hue sphere as computed using recently developed $N$-dimensional color theory for tetrachromacy~\cite{leeTetrachromacy2024}, customized to each of the observers here (i.e. with human Q or pigeon Q, respectively).  In the third dataset, the colors are sampled from an idealized tetrachromatic color space in which each of the cone channels is allowed to take any value, allowing theoretically maximum levels of chromatic contrast.  Since natural hyperspectral images have been found lacking in human tetrachromatic colors~\cite{leeTetrachromacy2024}, in principle we may expect the likelihood that 4D color vision emerges to increase across these three simulated environments.

Indeed, in line with theoretical intuition, Figure~\ref{fig:10} shows that 4D color vision emerges, as measured formally by \CMFSim, only for the third visual environment in the case of the human Q cone, but 4D color vision emerges with any of the visual environments for the pigeon Q cone. 
%50\% of women are reported to carry the genes for four distinct color photoreceptors~\cite{neitz2011genetics}, but we are societally unaware of tetrachromacy in the population with only one functional tetrachromat identified to date~\cite{jordan2010dimensionality}. One hypothesis is that 
Our simulations suggest strong genetic and environmental effects on emergence of tetrachromatic color vision.

\begin{figure}[t!]
    \centering
    \includegraphics[width=\columnwidth]{assets/fig10_0827.pdf}
    \caption{Results of expanded experiments of boosting color dimensionality from 3D to 4D, by addition of a Q cone to trichromatic L, M, S cones. Panel 1 models addition of a Q cone from natural human tetrachromacy, and panel 2 is for a Q cone from pigeon vision; spectral responses are shown.  Each panel compares emergence of color vision with three different visual environments: real hyperspectal images~\citep{arad2022ntire}; synthetic images containing tetrachromatic colors sampled from the hue sphere customized to that color observer~\citep{leeTetrachromacy2024}; synthetic images containing idealized tetrachromatic colors with chromatic contrast between cone channels at the theoretical maximum.  
    The simulation results show that 4D color vision only emerges for the human Q cone with the third environment, while it emerges for the pigeon Q cone under all three environments. }
    \label{fig:10}
\end{figure}

%% file: Rebuttal/RS-08-Extensive_Results.tex
\section{Further Evaluation of Color Emergence in Cortical Model}

\subsection{Validation of Simulated Color Matching Functions Against Human Psychophysical Data}\label{sec:supp-SandB}

Our simulation produces results that are highly consistent with the psychophysical measurements of actual human subjects. Specifically, we compared the output of \CMFSim with the empirical color matching function data reported in~\cite{stiles1955interim}. To match the experimental setup in~\cite{stiles1955interim}, we employed the same monochromatic color primaries at $444$nm, $526$nm, and $645$nm. As shown in Figure~\ref{fig:11}, the CMFs resulting from color matching simulation in \CMFSim closely align with the empirical color matching function curves, providing further validation of our simulated cortical model as an accurate representation of human trichromatic color vision.

\begin{figure*}[t!]
    \centering
    \includegraphics[width=0.72\textwidth]{assets/figX_1116.pdf}
    \caption{Side-by-side comparisons of two color-matching functions (CMFs) are presented. Left: The CMFs generated from our simulated cortical model, trained using a trichromatic retina. Right: The CMFs obtained from~\cite{stiles1955interim}, based on real human subjects. To ensure a direct comparison, our simulation used the same set of spectral primaries (i.e., 444nm, 526nm, and 645nm). This comparison highlights the validity of our trained cortical model as an accurate representation of a color observer.}
    \label{fig:11}
\end{figure*}

\subsection{Robust Emergence of Correct Color Dimensionality in Cortical Model}\label{sec:supp-consistency}

The emergence of correct color dimensionality in our simulations remained consistent across variations in initialization, designed to mimic the natural biological variability between humans~\citep{carroll2002estimates,hofer2005organization}. We tested this rigorously by varying both cone type ratios and initial noise parameters. First, we altered the L:M cone ratios in the retinal patch, testing 2:1 (original), 1:1, and 1:2, as well as an equal L:M:S ratio of 1:1:1. In all cases, the framework consistently converged to 3D color vision, demonstrating adaptability to different cone distributions. Second, we sampled the random seed for noise parameters, including: photon-shot noise during photoreceptor activation, lateral inhibition noise, Perlin noise for cell position randomization, and cortical model parameter initialization. Despite these changes, the model always converged to 3D color vision when trained with a trichromat retina, underscoring the robustness of the simulated cortical learning to biological and physical variability.

\begin{figure*}[t!]
    \centering
    \includegraphics[width=\textwidth]{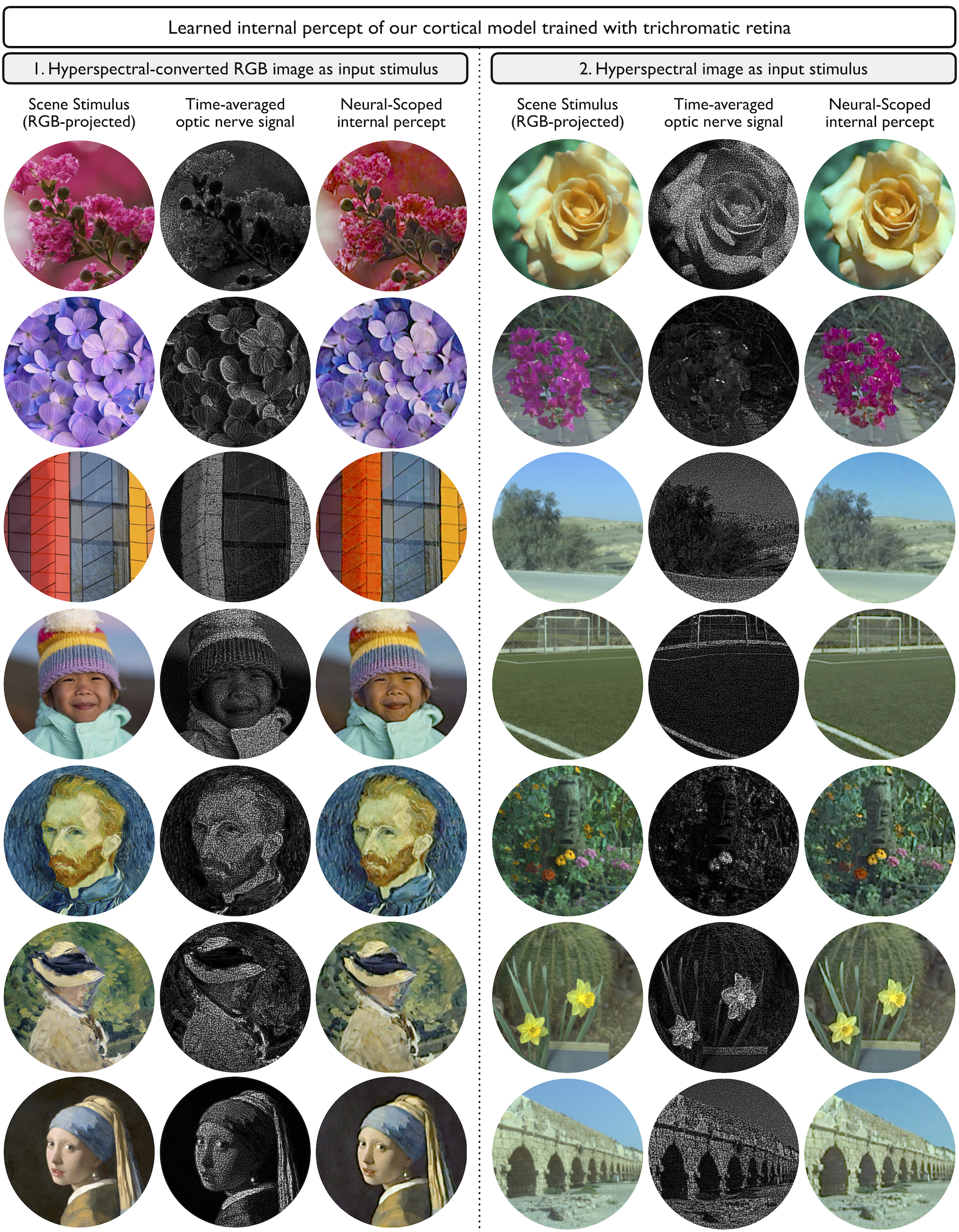}
    \caption{
    Our learned cortical model demonstrates robust performance across a wide variety of inputs during testing. Here, we present the neural-scoped internal percept of our cortical model trained with a trichromatic retina: 1. For hyperspectral images derived from RGB inputs (Section~\ref{sec:methods-scene-images}), the model accurately reconstructs the color percept (3rd column) from optic nerve signals (2nd column, time-averaged for clearer visualization). 
    2. For hyperspectral images~\citep{arad2022ntire}, the model similarly produces accurate color percepts (6th column) from optic nerve signals (5th column). For visual clarity, we show RGB-projected hyperspectral images in the 4th column.
    }
    \label{fig:14}
\end{figure*}

To further illustrate the robustness of our cortical model, we present its performance across a diverse range of input stimuli during testing, as shown in Figure~\ref{fig:14}. The figure demonstrates the Neural-Scoped internal percept of our cortical model trained with a trichromatic retina. Specifically:

\begin{enumerate}
\item For hyperspectral images derived from RGB inputs, converted using the method described in Section~\ref{sec:methods-scene-images}, the model successfully reconstructs accurate color percepts (3rd column) from optic nerve signals (2nd column), which have been time-averaged for improved visualization. Additionally, the model demonstrates complete robustness to various hue variations, accurately reconstructing color percepts even under significant changes in the chromatic properties of the input stimuli.
\item For hyperspectral images from a standard dataset~\citep{arad2022ntire}, the model accurately reconstructs color percepts (6th column) from optic nerve signals (5th column). To aid clarity, we include RGB-projected versions of the hyperspectral images in the 4th column.
\end{enumerate}

This highlights the model’s ability to generalize across varying input types and maintain robust performance under diverse conditions.

%% file: Rebuttal/RS-09-Table_of_Images.tex
\section{Table of Images with Varying Number of Cones for Photoreceptor Activations and RGC Spikes}

\begin{figure*}[t!]
    \centering
    \includegraphics[width=\textwidth]{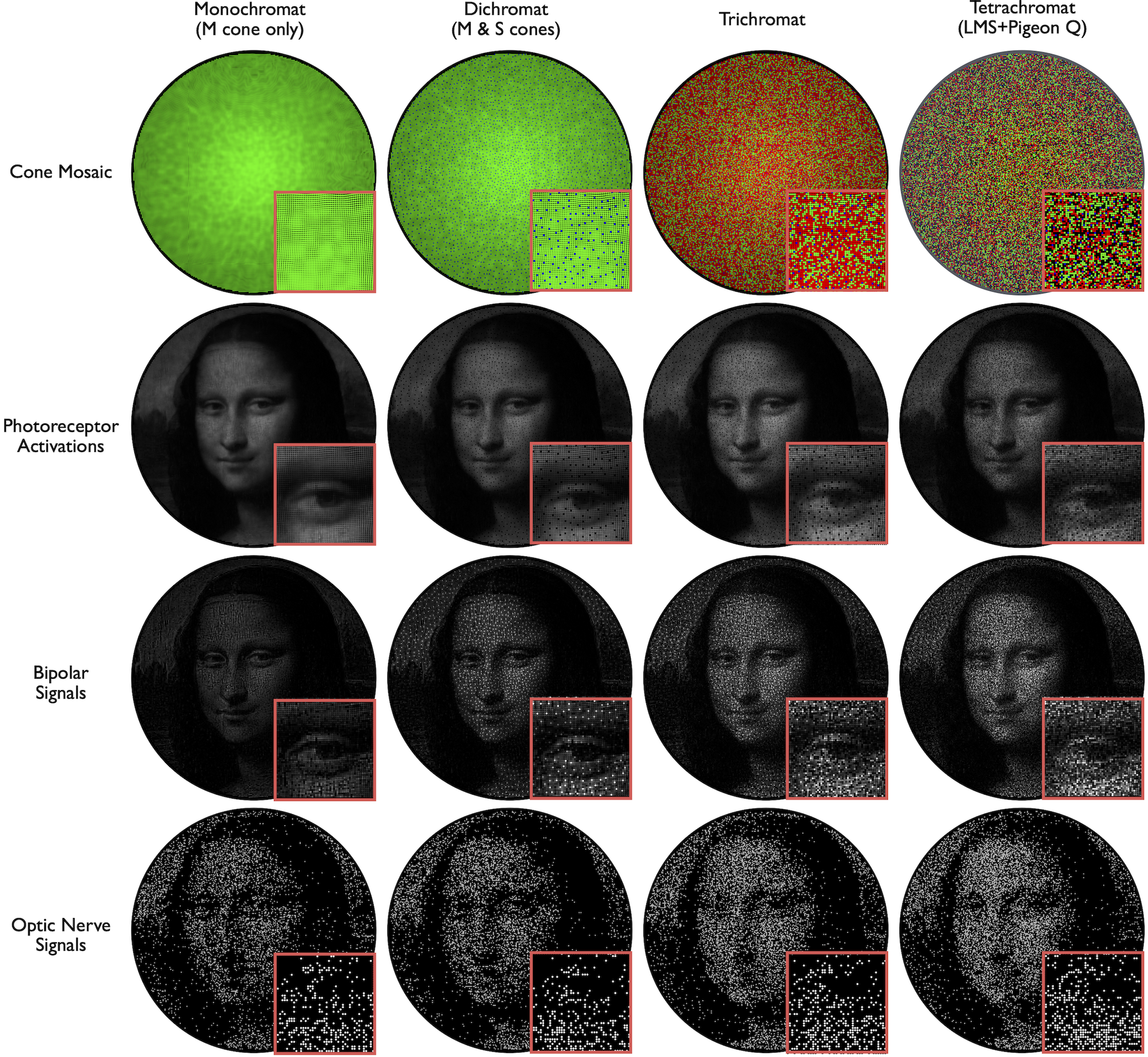}
    \caption{Expanded version of Figure~\ref{fig:2}.4, showing the full field of view with simulated retinal ganglion cell outputs (i.e., spiking optic nerve signals in the last row). Red boxes highlight close-up details of the full-sized data near the left eye of the Mona Lisa. L, M, S, and Q cones are visualized as red, green, blue, and black, respectively, in the first row.}
    \label{fig:12}
\end{figure*}

To enhance intuition and visual clarity, we present a table in Figure~\ref{fig:12}, where the number of cone types in the simulation varies across columns. The rows display images of photoreceptor activations, bipolar signals, and optic nerve signals.

The second row in Figure~\ref{fig:12} reveals that photoreceptor activations become increasingly noisy as additional cone types are added to the retinal mosaic. The third row shows bipolar signals, computed by applying a center-surround lateral inhibition kernel to the photoreceptor activations, introducing greater complexity compared to the photoreceptor layer. By the time optic nerve signals (spikes) are generated, differences between cone mosaics become almost indistinguishable, underscoring the cortical challenge of extracting color vision with the correct dimensionality.

\begin{figure*}[t!]
    \centering
    \includegraphics[width=\textwidth]{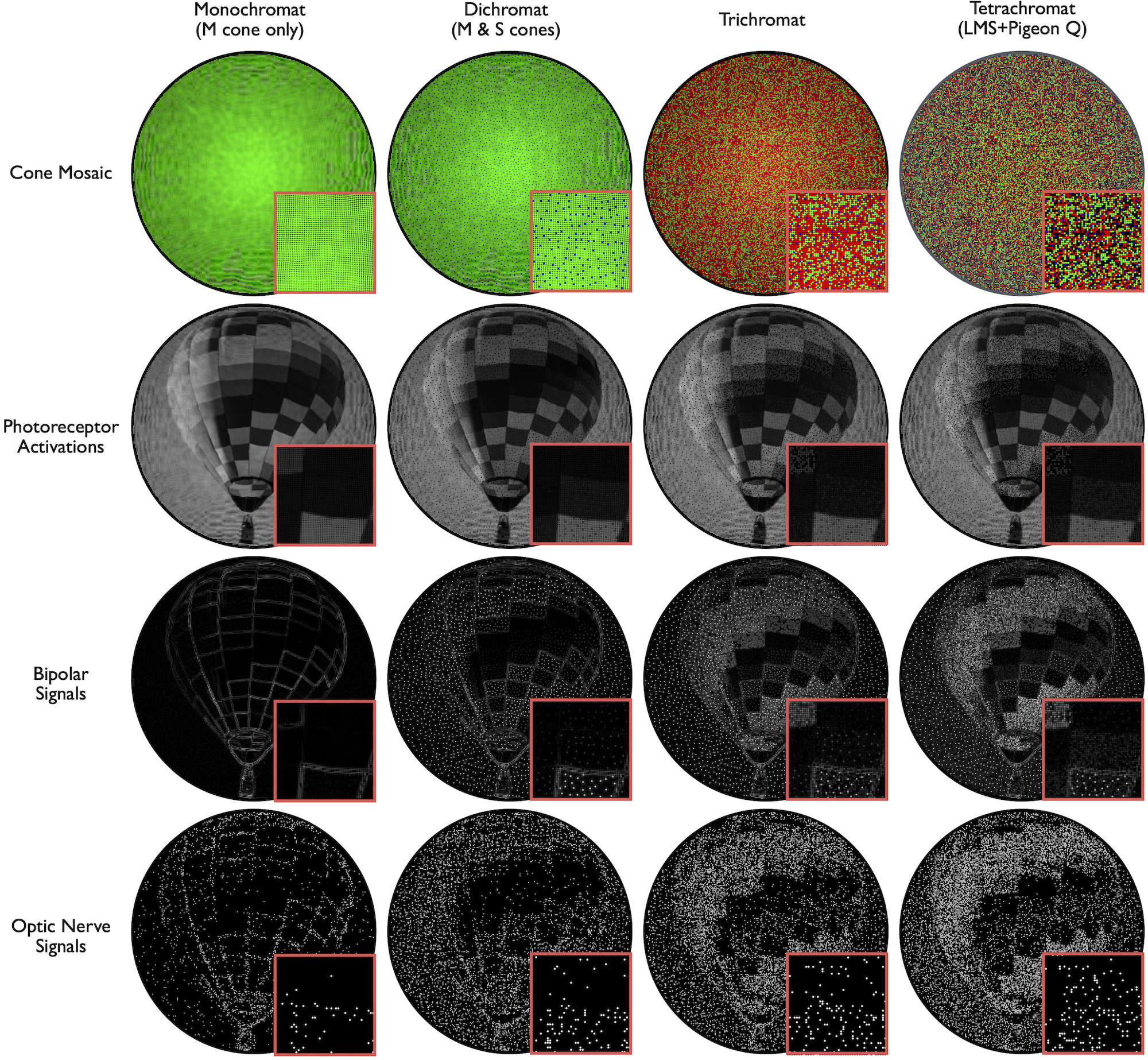}
    \caption{Expanded version of Figure~\ref{fig:2}.4, a variant of Figure~\ref{fig:12}, featuring the balloon image.}
    \label{fig:13}
\end{figure*}

We also present a variant of Figure~\ref{fig:12} that features the balloon image in Figure~\ref{fig:13}.

%% file: Rebuttal/RS-10-NS.tex
% \section{\red{a}}

% \setcounter{figure}{11}
% \begin{figure*}[t!]
%     \centering
%     \includegraphics[width=\textwidth]{assets/fig14_1123_2.pdf}
%     \caption{
%     Our learned cortical model demonstrates robust performance across a wide variety of inputs during testing. Here, we present the neural-scoped internal percept of our cortical model trained with a trichromatic retina: 1. For hyperspectral images derived from RGB inputs (Section~\ref{sec:methods-scene-images}), the model accurately reconstructs the color percept (3rd column) from optic nerve signals (2nd column, time-averaged for clearer visualization). 
%     2. For hyperspectral images~\citep{arad2022ntire}, the model similarly produces accurate color percepts (6th column) from optic nerve signals (5th column). For visual clarity, we show RGB-projected hyperspectral images in the 4th column.
%     }
%     \label{fig:14}
% \end{figure*}

% \newpage

% \begin{figure*}[t!]
%     \centering
%     \includegraphics[width=\textwidth]{assets/fig15_1124_2.pdf}
%     \caption{
%     A variant of Fig.~\ref{fig:14}, showcasing the robustness of our learned cortical model across a wide variety of input images, despite minimal hue diversity.
%     }
%     \label{fig:15}
% \end{figure*}

%% file: main.bbl
\begin{thebibliography}{178}
\providecommand{\natexlab}[1]{#1}
\providecommand{\url}[1]{\texttt{#1}}
\expandafter\ifx\csname urlstyle\endcsname\relax
  \providecommand{\doi}[1]{doi: #1}\else
  \providecommand{\doi}{doi: \begingroup \urlstyle{rm}\Url}\fi

\bibitem[CIE(1976)]{CIE1976}
\emph{Colorimetry}.
\newblock Number 15.2 in CIE Publication. Commission Internationale de l'Éclairage, Vienna, Austria, 2nd edition, 1976.

\bibitem[Abbott(1999)]{abbott1999lapicque}
Larry~F Abbott.
\newblock Lapicque’s introduction of the integrate-and-fire model neuron (1907).
\newblock \emph{Brain research bulletin}, 50\penalty0 (5-6):\penalty0 303--304, 1999.

\bibitem[Adams et~al.(1986)Adams, Maurer, and Davis]{adams1986newborns}
Russell~J Adams, Daphne Maurer, and Margaret Davis.
\newblock Newborns' discrimination of chromatic from achromatic stimuli.
\newblock \emph{Journal of Experimental Child Psychology}, 41\penalty0 (2):\penalty0 267--281, 1986.

\bibitem[Agustsson \& Timofte(2017)Agustsson and Timofte]{Agustsson_2017_CVPR_Workshops}
Eirikur Agustsson and Radu Timofte.
\newblock Ntire 2017 challenge on single image super-resolution: Dataset and study.
\newblock In \emph{The IEEE Conference on Computer Vision and Pattern Recognition (CVPR) Workshops}, July 2017.

\bibitem[Ahumada(1991)]{ahumada1991learning}
Albert~J Ahumada.
\newblock Learning receptor positions.
\newblock \emph{Computational models of visual processing}, pp.\  23--34, 1991.

\bibitem[Alpern et~al.(1983)Alpern, Kitahara, and Krantz]{alpern1983trit}
M~Alpern, K~Kitahara, and D~H Krantz.
\newblock Perception of colour in unilateral tritanopia.
\newblock \emph{The Journal of Physiology}, 335\penalty0 (1):\penalty0 683--697, 1983.

\bibitem[Anderson et~al.(2020)Anderson, Ratnam, Roorda, and Olshausen]{anderson2020high}
Alexander~G Anderson, Kavitha Ratnam, Austin Roorda, and Bruno~A Olshausen.
\newblock High-acuity vision from retinal image motion.
\newblock \emph{Journal of vision}, 20\penalty0 (7):\penalty0 34--34, 2020.

\bibitem[Angueyra et~al.(2022)Angueyra, Baudin, Schwartz, and Rieke]{angueyra2022predicting}
Juan~M Angueyra, Jacob Baudin, Gregory~W Schwartz, and Fred Rieke.
\newblock Predicting and manipulating cone responses to naturalistic inputs.
\newblock \emph{Journal of Neuroscience}, 42\penalty0 (7):\penalty0 1254--1274, 2022.

\bibitem[Arad et~al.(2022)Arad, Timofte, Yahel, Morag, Bernat, Cai, Lin, Lin, Wang, Zhang, et~al.]{arad2022ntire}
Boaz Arad, Radu Timofte, Rony Yahel, Nimrod Morag, Amir Bernat, Yuanhao Cai, Jing Lin, Zudi Lin, Haoqian Wang, Yulun Zhang, et~al.
\newblock Ntire 2022 spectral recovery challenge and data set.
\newblock In \emph{Proceedings of the IEEE/CVF Conference on Computer Vision and Pattern Recognition}, pp.\  863--881, 2022.

\bibitem[Araya \& Provis(1992)Araya and Provis]{araya1992evidence}
Claudia~Diaz Araya and Jan~M Provis.
\newblock Evidence of photoreceptor migration during early foveal development: a quantitative analysis of human fetal retinae.
\newblock \emph{Visual neuroscience}, 8\penalty0 (6):\penalty0 505--514, 1992.

\bibitem[Arbib(2003)]{arbib2003handbook}
Michael~A Arbib.
\newblock \emph{The handbook of brain theory and neural networks}.
\newblock MIT press, 2003.

\bibitem[Baradad et~al.(2021)Baradad, Wulff, Wang, Isola, and Torralba]{baradad2021learning}
Manel Baradad, Jonas Wulff, Tongzhou Wang, Phillip Isola, and Antonio Torralba.
\newblock Learning to see by looking at noise.
\newblock In \emph{Advances in Neural Information Processing Systems}, 2021.

\bibitem[Barlow et~al.(1961)]{barlow1961possible}
Horace~B Barlow et~al.
\newblock Possible principles underlying the transformation of sensory messages.
\newblock \emph{Sensory communication}, 1\penalty0 (01):\penalty0 217--233, 1961.

\bibitem[Bayer(1976)]{bayer1976color}
Bryce Bayer.
\newblock Color imaging array.
\newblock \emph{United States Patent, no. 3971065}, 1976.

\bibitem[Benson et~al.(2014)Benson, Manning, and Brainard]{benson2014unsupervised}
Noah~C Benson, Jeremy~R Manning, and David~H Brainard.
\newblock Unsupervised learning of cone spectral classes from natural images.
\newblock \emph{PLoS Computational Biology}, 10\penalty0 (6):\penalty0 e1003652, 2014.

\bibitem[Berry et~al.(1997)Berry, Warland, and Meister]{berry1997structure}
Michael~J Berry, David~K Warland, and Markus Meister.
\newblock The structure and precision of retinal spike trains.
\newblock \emph{Proceedings of the National Academy of Sciences}, 94\penalty0 (10):\penalty0 5411--5416, 1997.

\bibitem[Botella-Soler et~al.(2018)Botella-Soler, Deny, Martius, Marre, and Tka{\v{c}}ik]{botella2018nonlinear}
Vicente Botella-Soler, St{\'e}phane Deny, Georg Martius, Olivier Marre, and Ga{\v{s}}per Tka{\v{c}}ik.
\newblock Nonlinear decoding of a complex movie from the mammalian retina.
\newblock \emph{PLoS computational biology}, 14\penalty0 (5):\penalty0 e1006057, 2018.

\bibitem[Brackbill et~al.(2020)Brackbill, Rhoades, Kling, Shah, Sher, Litke, and Chichilnisky]{brackbill2020reconstruction}
Nora Brackbill, Colleen Rhoades, Alexandra Kling, Nishal~P Shah, Alexander Sher, Alan~M Litke, and EJ~Chichilnisky.
\newblock Reconstruction of natural images from responses of primate retinal ganglion cells.
\newblock \emph{Elife}, 9:\penalty0 e58516, 2020.

\bibitem[Brainard et~al.(2015)Brainard, Jiang, Cottaris, Rieke, Chichilnisky, Farrell, and Wandell]{brainard2015isetbio}
David Brainard, Haomiao Jiang, Nicolas~P Cottaris, Fred Rieke, EJ~Chichilnisky, Joyce~E Farrell, and Brian~A Wandell.
\newblock Isetbio: Computational tools for modeling early human vision.
\newblock In \emph{Imaging Systems and Applications}, pp.\  IT4A--4. Optica Publishing Group, 2015.

\bibitem[Brainard(2015)]{brainard2015color}
David~H Brainard.
\newblock Color and the cone mosaic.
\newblock \emph{Annual Review of Vision Science}, 1:\penalty0 519--546, 2015.

\bibitem[Brainard et~al.(2008)Brainard, Williams, and Hofer]{brainard2008trichromatic}
David~H Brainard, David~R Williams, and Heidi Hofer.
\newblock Trichromatic reconstruction from the interleaved cone mosaic: Bayesian model and the color appearance of small spots.
\newblock \emph{Journal of vision}, 8\penalty0 (5):\penalty0 15--15, 2008.

\bibitem[Brettel et~al.(1997)Brettel, Vi{\'e}not, and Mollon]{brettel1997computerized}
Hans Brettel, Fran{\c{c}}oise Vi{\'e}not, and John~D Mollon.
\newblock Computerized simulation of color appearance for dichromats.
\newblock \emph{Josa a}, 14\penalty0 (10):\penalty0 2647--2655, 1997.

\bibitem[Brouwer \& Heeger(2009)Brouwer and Heeger]{brouwer2009decoding}
Gijs~Joost Brouwer and David~J Heeger.
\newblock Decoding and reconstructing color from responses in human visual cortex.
\newblock \emph{Journal of Neuroscience}, 29\penalty0 (44):\penalty0 13992--14003, 2009.

\bibitem[Brown(1990)]{brown1990development}
Angela~M Brown.
\newblock Development of visual sensitivity to light and color vision in human infants: A critical review.
\newblock \emph{Vision Research}, 30\penalty0 (8):\penalty0 1159--1188, 1990.

\bibitem[Burak et~al.(2010)Burak, Rokni, Meister, and Sompolinsky]{burak2010bayesian}
Yoram Burak, Uri Rokni, Markus Meister, and Haim Sompolinsky.
\newblock Bayesian model of dynamic image stabilization in the visual system.
\newblock \emph{Proceedings of the National Academy of Sciences}, 107\penalty0 (45):\penalty0 19525--19530, 2010.

\bibitem[Carroll et~al.(2000)Carroll, McMahon, Neitz, and Neitz]{carroll2000flicker}
Joseph Carroll, Carrie McMahon, Maureen Neitz, and Jay Neitz.
\newblock Flicker-photometric electroretinogram estimates of l: M cone photoreceptor ratio in men with photopigment spectra derived from genetics.
\newblock \emph{JOSA A}, 17\penalty0 (3):\penalty0 499--509, 2000.

\bibitem[Carroll et~al.(2002)Carroll, Neitz, and Neitz]{carroll2002estimates}
Joseph Carroll, Jay Neitz, and Maureen Neitz.
\newblock Estimates of l: M cone ratio from erg flicker photometry and genetics.
\newblock \emph{Journal of vision}, 2\penalty0 (8):\penalty0 1--1, 2002.

\bibitem[Chen et~al.(2020)Chen, Kornblith, Norouzi, and Hinton]{chen2020simple}
Ting Chen, Simon Kornblith, Mohammad Norouzi, and Geoffrey Hinton.
\newblock A simple framework for contrastive learning of visual representations.
\newblock In \emph{International conference on machine learning}, pp.\  1597--1607. PMLR, 2020.

\bibitem[CIE(1931)]{commission1931commission}
CIE.
\newblock Commission internationale de l’eclairage proceedings, 1931.

\bibitem[Conway(2018)]{conway2018organization}
Bevil~R Conway.
\newblock The organization and operation of inferior temporal cortex.
\newblock \emph{Annual review of vision science}, 4:\penalty0 381--402, 2018.

\bibitem[Conway et~al.(2023)Conway, Malik-Moraleda, and Gibson]{conway2023color}
Bevil~R Conway, Saima Malik-Moraleda, and Edward Gibson.
\newblock Color appearance and the end of hering’s opponent-colors theory.
\newblock \emph{Trends in Cognitive Sciences}, 2023.

\bibitem[Cornelissen \& Brenner(2015)Cornelissen and Brenner]{cornelissen2015adding}
Frans~W Cornelissen and Eli Brenner.
\newblock Is adding a new class of cones to the retina sufficient to cure color-blindness?
\newblock \emph{Journal of vision}, 15\penalty0 (13):\penalty0 22--22, 2015.

\bibitem[Cottaris et~al.(2019)Cottaris, Jiang, Ding, Wandell, and Brainard]{cottaris2019computational}
Nicolas~P Cottaris, Haomiao Jiang, Xiaomao Ding, Brian~A Wandell, and David~H Brainard.
\newblock A computational-observer model of spatial contrast sensitivity: Effects of wave-front-based optics, cone-mosaic structure, and inference engine.
\newblock \emph{Journal of vision}, 19\penalty0 (4):\penalty0 8--8, 2019.

\bibitem[Croner \& Kaplan(1995)Croner and Kaplan]{croner1995receptive}
Lisa~J Croner and Ehud Kaplan.
\newblock Receptive fields of p and m ganglion cells across the primate retina.
\newblock \emph{Vision research}, 35\penalty0 (1):\penalty0 7--24, 1995.

\bibitem[Crook et~al.(2013)Crook, Packer, Troy, and Dacey]{crook2013synaptic}
Joanna~D Crook, Orin~S Packer, John~B Troy, and Dennis~M Dacey.
\newblock Synaptic mechanisms of color and luminance coding: rediscovering the xy-cell dichotomy in primate retinal ganglion cells.
\newblock In \emph{The new visual neurosciences}, pp.\  123--144. The MIT Press, Cambridge, MA, 2013.

\bibitem[Curcio et~al.(1990)Curcio, Sloan, Kalina, and Hendrickson]{curcio1990human}
Christine~A Curcio, Kenneth~R Sloan, Robert~E Kalina, and Anita~E Hendrickson.
\newblock Human photoreceptor topography.
\newblock \emph{Journal of comparative neurology}, 292\penalty0 (4):\penalty0 497--523, 1990.

\bibitem[Dacey et~al.(2000)Dacey, Packer, Diller, Brainard, Peterson, and Lee]{dacey2000center}
Dennis Dacey, Orin~S Packer, Lisa Diller, David Brainard, Beth Peterson, and Barry Lee.
\newblock Center surround receptive field structure of cone bipolar cells in primate retina.
\newblock \emph{Vision research}, 40\penalty0 (14):\penalty0 1801--1811, 2000.

\bibitem[Dacey(1993)]{dacey1993mosaic}
Dennis~M Dacey.
\newblock The mosaic of midget ganglion cells in the human retina.
\newblock \emph{Journal of Neuroscience}, 13\penalty0 (12):\penalty0 5334--5355, 1993.

\bibitem[Dacey \& Lee(1994)Dacey and Lee]{dacey1994blue}
Dennis~M Dacey and Barry~B Lee.
\newblock The'blue-on'opponent pathway in primate retina originates from a distinct bistratified ganglion cell type.
\newblock \emph{Nature}, 367\penalty0 (6465):\penalty0 731--735, 1994.

\bibitem[Dacey et~al.(1996)Dacey, Lee, Stafford, Pokorny, and Smith]{dacey1996horizontal}
Dennis~M Dacey, Barry~B Lee, Donna~K Stafford, Joel Pokorny, and Vivianne~C Smith.
\newblock Horizontal cells of the primate retina: cone specificity without spectral opponency.
\newblock \emph{Science}, 271\penalty0 (5249):\penalty0 656--659, 1996.

\bibitem[De~Sa(1993)]{de1993learning}
Virginia De~Sa.
\newblock Learning classification with unlabeled data.
\newblock \emph{Advances in neural information processing systems}, 6, 1993.

\bibitem[De~Valois \& De~Valois(1993)De~Valois and De~Valois]{de1993multi}
Russell~L De~Valois and Karen~K De~Valois.
\newblock A multi-stage color model.
\newblock \emph{Vision research}, 33\penalty0 (8):\penalty0 1053--1065, 1993.

\bibitem[Derrington et~al.(1984)Derrington, Krauskopf, and Lennie]{derrington1984chromatic}
Andrew~M Derrington, John Krauskopf, and Peter Lennie.
\newblock Chromatic mechanisms in lateral geniculate nucleus of macaque.
\newblock \emph{The Journal of physiology}, 357\penalty0 (1):\penalty0 241--265, 1984.

\bibitem[Descartes(1988)]{descartes1988descartes}
Ren{\'e} Descartes.
\newblock \emph{Descartes: Selected philosophical writings}.
\newblock Cambridge University Press, 1988.

\bibitem[Dinh et~al.(2016)Dinh, Sohl-Dickstein, and Bengio]{dinh2016density}
Laurent Dinh, Jascha Sohl-Dickstein, and Samy Bengio.
\newblock Density estimation using real nvp.
\newblock \emph{arXiv preprint arXiv:1605.08803}, 2016.

\bibitem[Dougherty et~al.(2003)Dougherty, Koch, Brewer, Fischer, Modersitzki, and Wandell]{dougherty2003visual}
Robert~F Dougherty, Volker~M Koch, Alyssa~A Brewer, Bernd Fischer, Jan Modersitzki, and Brian~A Wandell.
\newblock Visual field representations and locations of visual areas v1/2/3 in human visual cortex.
\newblock \emph{Journal of vision}, 3\penalty0 (10):\penalty0 1--1, 2003.

\bibitem[Dowling \& Boycott(1966)Dowling and Boycott]{dowling1966organization}
John~E Dowling and Brian~Blundell Boycott.
\newblock Organization of the primate retina: electron microscopy.
\newblock \emph{Proceedings of the Royal Society of London. Series B. Biological Sciences}, 166\penalty0 (1002):\penalty0 80--111, 1966.

\bibitem[Enroth-Cugell et~al.(1983)Enroth-Cugell, Robson, Schweitzer-Tong, and Watson]{enroth1983spatio}
Ch~Enroth-Cugell, JG~Robson, DE~Schweitzer-Tong, and AB~Watson.
\newblock Spatio-temporal interactions in cat retinal ganglion cells showing linear spatial summation.
\newblock \emph{The Journal of Physiology}, 341\penalty0 (1):\penalty0 279--307, 1983.

\bibitem[Fairchild(2013)]{fairchild2013color}
Mark~D Fairchild.
\newblock \emph{Color appearance models}.
\newblock John Wiley \& Sons, 2013.

\bibitem[Field et~al.(2010)Field, Gauthier, Sher, Greschner, Machado, Jepson, Shlens, Gunning, Mathieson, Dabrowski, et~al.]{field2010functional}
Greg~D Field, Jeffrey~L Gauthier, Alexander Sher, Martin Greschner, Timothy~A Machado, Lauren~H Jepson, Jonathon Shlens, Deborah~E Gunning, Keith Mathieson, Wladyslaw Dabrowski, et~al.
\newblock Functional connectivity in the retina at the resolution of photoreceptors.
\newblock \emph{Nature}, 467\penalty0 (7316):\penalty0 673--677, 2010.

\bibitem[F{\"o}ldi{\'a}k(1991)]{foldiak1991learning}
Peter F{\"o}ldi{\'a}k.
\newblock Learning invariance from transformation sequences.
\newblock \emph{Neural computation}, 3\penalty0 (2):\penalty0 194--200, 1991.

\bibitem[Fossum(1997)]{fossum1997cmos}
Eric~R Fossum.
\newblock Cmos image sensors: Electronic camera-on-a-chip.
\newblock \emph{IEEE transactions on electron devices}, 44\penalty0 (10):\penalty0 1689--1698, 1997.

\bibitem[Foster(2011)]{FOSTER2011674}
David~H. Foster.
\newblock Color constancy.
\newblock \emph{Vision Research}, 51\penalty0 (7):\penalty0 674--700, 2011.
\newblock ISSN 0042-6989.
\newblock \doi{https://doi.org/10.1016/j.visres.2010.09.006}.
\newblock Vision Research 50th Anniversary Issue: Part 1.

\bibitem[Freedland \& Rieke(2022)Freedland and Rieke]{freedland2022systematic}
Julian Freedland and Fred Rieke.
\newblock Systematic reduction of the dimensionality of natural scenes allows accurate predictions of retinal ganglion cell spike outputs.
\newblock \emph{Proceedings of the National Academy of Sciences}, 119\penalty0 (46):\penalty0 e2121744119, 2022.

\bibitem[Godat et~al.(2022)Godat, Cottaris, Patterson, Kohout, Parkins, Yang, Strazzeri, McGregor, Brainard, Merigan, et~al.]{godat2022vivo}
Tyler Godat, Nicolas~P Cottaris, Sara Patterson, Kendall Kohout, Keith Parkins, Qiang Yang, Jennifer~M Strazzeri, Juliette~E McGregor, David~H Brainard, William~H Merigan, et~al.
\newblock In vivo chromatic and spatial tuning of foveolar retinal ganglion cells in macaca fascicularis.
\newblock \emph{Plos one}, 17\penalty0 (11):\penalty0 e0278261, 2022.

\bibitem[Graham \& Hsia(1959)Graham and Hsia]{graham1959studies}
CH~Graham and Yun Hsia.
\newblock Studies of color blindness: a unilaterally dichromatic subject, 1959.

\bibitem[Grassmann(1853)]{grassmann1853theorie}
Hermann Grassmann.
\newblock Zur theorie der farbenmischung.
\newblock \emph{Annalen der Physik}, 165\penalty0 (5):\penalty0 69--84, 1853.

\bibitem[Guild(1931)]{guild1931colorimetric}
John Guild.
\newblock The colorimetric properties of the spectrum.
\newblock \emph{Philosophical Transactions of the Royal Society of London. Series A, Containing Papers of a Mathematical or Physical Character}, 230\penalty0 (681-693):\penalty0 149--187, 1931.

\bibitem[Guth(1991)]{guth1991model}
S~Lee Guth.
\newblock Model for color vision and light adaptation.
\newblock \emph{JOSA A}, 8\penalty0 (6):\penalty0 976--993, 1991.

\bibitem[Haile et~al.(2019)Haile, Bohon, Romero, and Conway]{haile2019visual}
Theodros~M Haile, Kaitlin~S Bohon, Maria~C Romero, and Bevil~R Conway.
\newblock Visual stimulus-driven functional organization of macaque prefrontal cortex.
\newblock \emph{Neuroimage}, 188:\penalty0 427--444, 2019.

\bibitem[Hartline \& Ratliff(1958)Hartline and Ratliff]{hartline1958spatial}
H~KEFFER Hartline and Floyd Ratliff.
\newblock Spatial summation of inhibitory influences in the eye of limulus, and the mutual interaction of receptor units.
\newblock \emph{The Journal of general physiology}, 41\penalty0 (5):\penalty0 1049--1066, 1958.

\bibitem[He et~al.(2022)He, Chen, Xie, Li, Doll{\'a}r, and Girshick]{he2022masked}
Kaiming He, Xinlei Chen, Saining Xie, Yanghao Li, Piotr Doll{\'a}r, and Ross Girshick.
\newblock Masked autoencoders are scalable vision learners.
\newblock In \emph{Proceedings of the IEEE/CVF conference on computer vision and pattern recognition}, pp.\  16000--16009, 2022.

\bibitem[Hering(1878)]{hering1878lehre}
Ewald Hering.
\newblock \emph{Zur Lehre vom Lichtsinne: sechs Mittheilungen an die Kaiserl. Akademie der Wissenschaften in Wien}.
\newblock C. Gerold's Sohn, 1878.

\bibitem[Hinton(2022)]{hinton2022forward}
Geoffrey Hinton.
\newblock The forward-forward algorithm: Some preliminary investigations.
\newblock \emph{arXiv preprint arXiv:2212.13345}, 2022.

\bibitem[Hinton(1990)]{hinton1990connectionist}
Geoffrey~E Hinton.
\newblock Connectionist learning procedures.
\newblock In \emph{Machine learning}, pp.\  555--610. Elsevier, 1990.

\bibitem[Hirasawa \& Kaneko(2003)Hirasawa and Kaneko]{hirasawa2003pH}
Hajime Hirasawa and Akimichi Kaneko.
\newblock {pH Changes in the Invaginating Synaptic Cleft Mediate Feedback from Horizontal Cells to Cone Photoreceptors by Modulating Ca2+ Channels }.
\newblock \emph{Journal of General Physiology}, 122\penalty0 (6):\penalty0 657--671, 11 2003.
\newblock ISSN 0022-1295.

\bibitem[Hofer et~al.(2005)Hofer, Carroll, Neitz, Neitz, and Williams]{hofer2005organization}
Heidi Hofer, Joseph Carroll, Jay Neitz, Maureen Neitz, and David~R Williams.
\newblock Organization of the human trichromatic cone mosaic.
\newblock \emph{Journal of Neuroscience}, 25\penalty0 (42):\penalty0 9669--9679, 2005.

\bibitem[Holmes(1918)]{holmes1918disturbances}
Gordon Holmes.
\newblock Disturbances of vision by cerebral lesions.
\newblock \emph{The British journal of ophthalmology}, 2\penalty0 (7):\penalty0 353, 1918.

\bibitem[Hubel \& Wiesel(1962)Hubel and Wiesel]{hubel1962receptive}
David~H Hubel and Torsten~N Wiesel.
\newblock Receptive fields, binocular interaction and functional architecture in the cat's visual cortex.
\newblock \emph{The Journal of physiology}, 160\penalty0 (1):\penalty0 106, 1962.

\bibitem[Jacobs(2018)]{jacobs2018photopigments}
Gerald~H Jacobs.
\newblock Photopigments and the dimensionality of animal color vision.
\newblock \emph{Neuroscience \& Biobehavioral Reviews}, 86:\penalty0 108--130, 2018.

\bibitem[Jacobs et~al.(2007)Jacobs, Williams, Cahill, and Nathans]{jacobs2007emergence}
Gerald~H Jacobs, Gary~A Williams, Hugh Cahill, and Jeremy Nathans.
\newblock Emergence of novel color vision in mice engineered to express a human cone photopigment.
\newblock \emph{science}, 315\penalty0 (5819):\penalty0 1723--1725, 2007.

\bibitem[Jim{\'e}nez \& Malo(2013)Jim{\'e}nez and Malo]{jimenez2013role}
Sandra Jim{\'e}nez and Jes{\'u}s Malo.
\newblock The role of spatial information in disentangling the irradiance--reflectance--transmittance ambiguity.
\newblock \emph{IEEE transactions on geoscience and remote sensing}, 52\penalty0 (8):\penalty0 4881--4894, 2013.

\bibitem[Jordan et~al.(2010)Jordan, Deeb, Bosten, and Mollon]{jordan2010dimensionality}
Gabriele Jordan, Samir~S Deeb, Jenny~M Bosten, and John~D Mollon.
\newblock The dimensionality of color vision in carriers of anomalous trichromacy.
\newblock \emph{Journal of vision}, 10\penalty0 (8):\penalty0 12--12, 2010.

\bibitem[Judd(1948)]{judd1948color}
Deane~B Judd.
\newblock Color perceptions of deuteranopic and protanopic observers.
\newblock \emph{Journal of Research of the National Bureau of Standards}, 41:\penalty0 247--271, 1948.

\bibitem[Judd(1949)]{judd1949response}
Deane~B Judd.
\newblock \emph{Response functions for types of vision according to the M{\"u}ller theory}.
\newblock US Government Printing Office, 1949.

\bibitem[Kim et~al.(2021)Kim, Brackbill, Batty, Lee, Mitelut, Tong, Chichilnisky, and Paninski]{kim2021nonlinear}
Young~Joon Kim, Nora Brackbill, Eleanor Batty, JinHyung Lee, Catalin Mitelut, William Tong, EJ~Chichilnisky, and Liam Paninski.
\newblock Nonlinear decoding of natural images from large-scale primate retinal ganglion recordings.
\newblock \emph{Neural Computation}, 33\penalty0 (7):\penalty0 1719--1750, 2021.

\bibitem[Kimmel(1999)]{kimmel1999demosaicing}
Ron Kimmel.
\newblock Demosaicing: Image reconstruction from color ccd samples.
\newblock \emph{IEEE Transactions on image processing}, 8\penalty0 (9):\penalty0 1221--1228, 1999.

\bibitem[Kingma \& Ba(2014)Kingma and Ba]{kingma2014adam}
Diederik~P Kingma and Jimmy Ba.
\newblock Adam: A method for stochastic optimization.
\newblock \emph{arXiv preprint arXiv:1412.6980}, 2014.

\bibitem[Krauskopf \& Karl(1992)Krauskopf and Karl]{krauskopf1992color}
John Krauskopf and Gegenfurtner Karl.
\newblock Color discrimination and adaptation.
\newblock \emph{Vision research}, 32\penalty0 (11):\penalty0 2165--2175, 1992.

\bibitem[Kuffler(1953)]{kuffler1953discharge}
Stephen~W Kuffler.
\newblock Discharge patterns and functional organization of mammalian retina.
\newblock \emph{Journal of neurophysiology}, 16\penalty0 (1):\penalty0 37--68, 1953.

\bibitem[Lafer-Sousa \& Conway(2013)Lafer-Sousa and Conway]{lafer2013parallel}
Rosa Lafer-Sousa and Bevil~R Conway.
\newblock Parallel, multi-stage processing of colors, faces and shapes in macaque inferior temporal cortex.
\newblock \emph{Nature neuroscience}, 16\penalty0 (12):\penalty0 1870--1878, 2013.

\bibitem[Land(1977)]{land1977retinex}
Edwin~H Land.
\newblock The retinex theory of color vision.
\newblock \emph{Scientific american}, 237\penalty0 (6):\penalty0 108--129, 1977.

\bibitem[Land \& Nilsson(2012)Land and Nilsson]{land2012animal}
Michael~F Land and Dan-Eric Nilsson.
\newblock \emph{Animal eyes}.
\newblock OUP Oxford, 2012.

\bibitem[Lapicque(1907)]{lapicque1907recherches}
L~Lapicque.
\newblock Recherches quantitatives sur l’excitation electrique des nerfs.
\newblock \emph{J. Physiol. Paris}, 9:\penalty0 620--635, 1907.

\bibitem[Lee et~al.(2024)Lee, Jennings, Srivastava, and Ng]{leeTetrachromacy2024}
Jessica Lee, Nicholas Jennings, Varun Srivastava, and Ren Ng.
\newblock Theory of human tetrachromatic color experience and printing.
\newblock \emph{ACM Trans. Graph}, 43\penalty0 (4), July 2024.

\bibitem[Lennie et~al.(1991)Lennie, Haake, and Williams]{lennie1991design}
Peter Lennie, P~William Haake, and David~R Williams.
\newblock The design of chromatically opponent receptive fields.
\newblock \emph{Computational models of visual processing}, pp.\  71--82, 1991.

\bibitem[Li et~al.(2014)Li, Liu, Juusola, and Tang]{li2014perceptual}
Ming Li, Fang Liu, Mikko Juusola, and Shiming Tang.
\newblock Perceptual color map in macaque visual area v4.
\newblock \emph{Journal of Neuroscience}, 34\penalty0 (1):\penalty0 202--217, 2014.

\bibitem[Lian et~al.(2019)Lian, MacKenzie, Brainard, Cottaris, and Wandell]{lian2019ray}
Trisha Lian, Kevin~J MacKenzie, David~H Brainard, Nicolas~P Cottaris, and Brian~A Wandell.
\newblock Ray tracing 3d spectral scenes through human optics models.
\newblock \emph{Journal of vision}, 19\penalty0 (12):\penalty0 23--23, 2019.

\bibitem[Lillicrap et~al.(2020)Lillicrap, Santoro, Marris, Akerman, and Hinton]{lillicrap2020backpropagation}
Timothy~P Lillicrap, Adam Santoro, Luke Marris, Colin~J Akerman, and Geoffrey Hinton.
\newblock Backpropagation and the brain.
\newblock \emph{Nature Reviews Neuroscience}, 21\penalty0 (6):\penalty0 335--346, 2020.

\bibitem[Litke et~al.(2004)Litke, Bezayiff, Chichilnisky, Cunningham, Dabrowski, Grillo, Grivich, Grybos, Hottowy, Kachiguine, et~al.]{litke2004does}
AM~Litke, N~Bezayiff, EJ~Chichilnisky, W~Cunningham, W~Dabrowski, AA~Grillo, M~Grivich, P~Grybos, P~Hottowy, S~Kachiguine, et~al.
\newblock What does the eye tell the brain?: Development of a system for the large-scale recording of retinal output activity.
\newblock \emph{IEEE Transactions on Nuclear Science}, 51\penalty0 (4):\penalty0 1434--1440, 2004.

\bibitem[Liu et~al.(2022)Liu, Karamanlis, and Gollisch]{liu2022simple}
Jian~K Liu, Dimokratis Karamanlis, and Tim Gollisch.
\newblock Simple model for encoding natural images by retinal ganglion cells with nonlinear spatial integration.
\newblock \emph{PLoS computational biology}, 18\penalty0 (3):\penalty0 e1009925, 2022.

\bibitem[Liu et~al.(2018)Liu, Jung, Dubra, and Tam]{liu2018cone}
Jianfei Liu, HaeWon Jung, Alfredo Dubra, and Johnny Tam.
\newblock Cone photoreceptor cell segmentation and diameter measurement on adaptive optics images using circularly constrained active contour model.
\newblock \emph{Investigative ophthalmology \& visual science}, 59\penalty0 (11):\penalty0 4639--4652, 2018.

\bibitem[Liu et~al.(2020)Liu, Li, Zhang, Lu, Gong, Yin, Chen, Qian, Yang, Andolina, et~al.]{liu2020hierarchical}
Ye~Liu, Ming Li, Xian Zhang, Yiliang Lu, Hongliang Gong, Jiapeng Yin, Zheyuan Chen, Liling Qian, Yupeng Yang, Ian~Max Andolina, et~al.
\newblock Hierarchical representation for chromatic processing across macaque v1, v2, and v4.
\newblock \emph{Neuron}, 108\penalty0 (3):\penalty0 538--550, 2020.

\bibitem[Livingstone \& Hubel(1987)Livingstone and Hubel]{livingstone1987psychophysical}
Margaret~S Livingstone and David~H Hubel.
\newblock Psychophysical evidence for separate channels for the perception of form, color, movement, and depth.
\newblock \emph{Journal of Neuroscience}, 7\penalty0 (11):\penalty0 3416--3468, 1987.

\bibitem[Lotter et~al.(2016)Lotter, Kreiman, and Cox]{lotter2016deep}
William Lotter, Gabriel Kreiman, and David Cox.
\newblock Deep predictive coding networks for video prediction and unsupervised learning.
\newblock In \emph{International Conference on Learning Representations}, 2016.

\bibitem[MacAdam(1942)]{macadam1942visual}
David~L MacAdam.
\newblock Visual sensitivities to color differences in daylight.
\newblock \emph{Josa}, 32\penalty0 (5):\penalty0 247--274, 1942.

\bibitem[MacLeod \& Boynton(1979)MacLeod and Boynton]{macleod1979chromaticity}
Donald~IA MacLeod and Robert~M Boynton.
\newblock Chromaticity diagram showing cone excitation by stimuli of equal luminance.
\newblock \emph{JOSA}, 69\penalty0 (8):\penalty0 1183--1186, 1979.

\bibitem[Makous(2007)]{makous2007comment}
Walter Makous.
\newblock Comment on" emergence of novel color vision in mice engineered to express a human cone photopigment".
\newblock \emph{Science}, 318\penalty0 (5848):\penalty0 196--196, 2007.

\bibitem[Maloney \& Ahumada(1989)Maloney and Ahumada]{maloney_ahumada}
Laurence~T Maloney and Albert~J Ahumada.
\newblock Learning by assertion: Two methods for calibrating a linear visual system.
\newblock \emph{Neural Computation}, 1\penalty0 (3):\penalty0 392--401, 1989.

\bibitem[Mancuso et~al.(2009)Mancuso, Hauswirth, Li, Connor, Kuchenbecker, Mauck, Neitz, and Neitz]{mancuso2009gene}
Katherine Mancuso, William~W Hauswirth, Qiuhong Li, Thomas~B Connor, James~A Kuchenbecker, Matthew~C Mauck, Jay Neitz, and Maureen Neitz.
\newblock Gene therapy for red--green colour blindness in adult primates.
\newblock \emph{Nature}, 461\penalty0 (7265):\penalty0 784--787, 2009.

\bibitem[Marre et~al.(2017)Marre, Tkacik, Amodei, Schneidman, Bialek, and Berry]{marre2017multi}
Olivier Marre, Gasper Tkacik, Dario Amodei, Elad Schneidman, William Bialek, and Michael Berry.
\newblock Multi-electrode array recording from salamander retinal ganglion cells.
\newblock 2017.

\bibitem[Martinez-Conde et~al.(2013)Martinez-Conde, Otero-Millan, and Macknik]{martinez2013impact}
Susana Martinez-Conde, Jorge Otero-Millan, and Stephen~L Macknik.
\newblock The impact of microsaccades on vision: towards a unified theory of saccadic function.
\newblock \emph{Nature Reviews Neuroscience}, 14\penalty0 (2):\penalty0 83--96, 2013.

\bibitem[Maxwell(1856)]{maxwell1856theory}
James~Clerk Maxwell.
\newblock Theory of the perception of colors.
\newblock \emph{Trans. R. Scottish Soc. Arts}, 4:\penalty0 394--400, 1856.

\bibitem[McClurkin \& Optican(1996)McClurkin and Optican]{mcclurkin1996primate1}
J.~W. McClurkin and L.~M. Optican.
\newblock Primate striate and prestriate cortical neurons during discrimination. i. simultaneous temporal encoding of information about color and pattern.
\newblock \emph{Journal of Neurophysiology}, 75\penalty0 (1):\penalty0 481--495, 1996.

\bibitem[McClurkin et~al.(1996)McClurkin, Zarbock, and Optican]{mcclurkin1996primate2}
J.~W. McClurkin, J.~A. Zarbock, and L.~M Optican.
\newblock Primate striate and prestriate cortical neurons during discrimination. ii. separable temporal codes for color and pattern.
\newblock \emph{Journal of Neurophysiology}, 75\penalty0 (1):\penalty0 496--507, 1996.

\bibitem[McLaren(1976)]{mclaren1976xiii}
K~McLaren.
\newblock Xiii—the development of the cie 1976 (l* a* b*) uniform colour space and colour-difference formula.
\newblock \emph{Journal of the Society of Dyers and Colourists}, 92\penalty0 (9):\penalty0 338--341, 1976.

\bibitem[McMahon et~al.(2000)McMahon, Lankheet, Lennie, and Williams]{mcmahon2000fine}
Matthew~J McMahon, Martin~JM Lankheet, Peter Lennie, and David~R Williams.
\newblock Fine structure of parvocellular receptive fields in the primate fovea revealed by laser interferometry.
\newblock \emph{Journal of Neuroscience}, 20\penalty0 (5):\penalty0 2043--2053, 2000.

\bibitem[Mollon \& Bowmaker(1992)Mollon and Bowmaker]{mollon1992spatial}
JD~Mollon and JK~Bowmaker.
\newblock The spatial arrangement of cones in the primate fovea.
\newblock \emph{Nature}, 360\penalty0 (6405):\penalty0 677--679, 1992.

\bibitem[Naselaris et~al.(2009)Naselaris, Prenger, Kay, Oliver, and Gallant]{naselaris2009bayesian}
Thomas Naselaris, Ryan~J Prenger, Kendrick~N Kay, Michael Oliver, and Jack~L Gallant.
\newblock Bayesian reconstruction of natural images from human brain activity.
\newblock \emph{Neuron}, 63\penalty0 (6):\penalty0 902--915, 2009.

\bibitem[Neitz \& Neitz(2017)Neitz and Neitz]{neitz2017evolution}
J~Neitz and M~Neitz.
\newblock Evolution of the circuitry for conscious color vision in primates.
\newblock \emph{Eye}, 31\penalty0 (2):\penalty0 286--300, 2017.

\bibitem[Neitz \& Neitz(2011)Neitz and Neitz]{neitz2011genetics}
Jay Neitz and Maureen Neitz.
\newblock The genetics of normal and defective color vision.
\newblock \emph{Vision Research}, 51\penalty0 (7):\penalty0 633--651, 2011.

\bibitem[Newton(1704)]{newton1704opticks}
Isaac Newton.
\newblock \emph{Opticks: or, A Treatise of the Reflexions, Refractions, Inflexions and Colours of Light}.
\newblock Courier Corporation, 1704.

\bibitem[Nishimoto et~al.(2011)Nishimoto, Vu, Naselaris, Benjamini, Yu, and Gallant]{nishimoto2011reconstructing}
Shinji Nishimoto, An~T Vu, Thomas Naselaris, Yuval Benjamini, Bin Yu, and Jack~L Gallant.
\newblock Reconstructing visual experiences from brain activity evoked by natural movies.
\newblock \emph{Current biology}, 21\penalty0 (19):\penalty0 1641--1646, 2011.

\bibitem[O'Regan \& No{\"e}(2001)O'Regan and No{\"e}]{o2001sensorimotor}
J~Kevin O'Regan and Alva No{\"e}.
\newblock A sensorimotor account of vision and visual consciousness.
\newblock \emph{Behavioral and brain sciences}, 24\penalty0 (5):\penalty0 939--973, 2001.

\bibitem[Osterberg(1935)]{osterberg1935topography}
Gustav~A Osterberg.
\newblock Topography of the layer of rods and cones in the human retina.
\newblock \emph{Acta ophthalmologica}, 1935.

\bibitem[Packer \& Dacey(2002)Packer and Dacey]{packer2002receptive}
Orin~S Packer and Dennis~M Dacey.
\newblock Receptive field structure of h1 horizontal cells in macaque monkey retina.
\newblock \emph{Journal of Vision}, 2\penalty0 (4):\penalty0 1--1, 2002.

\bibitem[Palmer et~al.(2015)Palmer, Marre, Berry, and Bialek]{palmer2015predictive}
Stephanie~E Palmer, Olivier Marre, Michael~J Berry, and William Bialek.
\newblock Predictive information in a sensory population.
\newblock \emph{Proceedings of the National Academy of Sciences}, 112\penalty0 (22):\penalty0 6908--6913, 2015.

\bibitem[Parthasarathy et~al.(2017)Parthasarathy, Batty, Falcon, Rutten, Rajpal, Chichilnisky, and Paninski]{parthasarathy2017neural}
Nikhil Parthasarathy, Eleanor Batty, William Falcon, Thomas Rutten, Mohit Rajpal, EJ~Chichilnisky, and Liam Paninski.
\newblock Neural networks for efficient bayesian decoding of natural images from retinal neurons.
\newblock \emph{Advances in Neural Information Processing Systems}, 30, 2017.

\bibitem[Perlin(1985)]{perlin1985image}
Ken Perlin.
\newblock An image synthesizer.
\newblock \emph{ACM Siggraph Computer Graphics}, 19\penalty0 (3):\penalty0 287--296, 1985.

\bibitem[Pinna(1987)]{pinna1987effetto}
Baingio Pinna.
\newblock Un effetto di colorazione.
\newblock In \emph{Il laboratorio e la citt{\`a}. XXI Congresso degli Psicologi Italiani}, volume 158. Edizioni SIPs, Societ{\'a} Italiana di Psiocologia, Milano, 1987.

\bibitem[Pinna et~al.(2001)Pinna, Brelstaff, and Spillmann]{pinna2001surface}
Baingio Pinna, Gavin Brelstaff, and Lothar Spillmann.
\newblock Surface color from boundaries: a new ‘watercolor’illusion.
\newblock \emph{Vision research}, 41\penalty0 (20):\penalty0 2669--2676, 2001.

\bibitem[Polimeni et~al.(2010)Polimeni, Fischl, Greve, and Wald]{polimeni2010laminar}
Jonathan~R Polimeni, Bruce Fischl, Douglas~N Greve, and Lawrence~L Wald.
\newblock Laminar analysis of 7 t bold using an imposed spatial activation pattern in human v1.
\newblock \emph{Neuroimage}, 52\penalty0 (4):\penalty0 1334--1346, 2010.

\bibitem[Ramanath et~al.(2002)Ramanath, Snyder, Bilbro, and Sander~III]{ramanath2002demosaicking}
Rajeev Ramanath, Wesley~E Snyder, Griff~L Bilbro, and William~A Sander~III.
\newblock Demosaicking methods for bayer color arrays.
\newblock \emph{Journal of Electronic imaging}, 11\penalty0 (3):\penalty0 306--315, 2002.

\bibitem[Rao \& Ballard(1999)Rao and Ballard]{rao1999predictive}
Rajesh~PN Rao and Dana~H Ballard.
\newblock Predictive coding in the visual cortex: a functional interpretation of some extra-classical receptive-field effects.
\newblock \emph{Nature neuroscience}, 2\penalty0 (1):\penalty0 79--87, 1999.

\bibitem[Reinhard \& M{\"u}nch(2021)Reinhard and M{\"u}nch]{reinhard2021visual}
Katja Reinhard and Thomas~A M{\"u}nch.
\newblock Visual properties of human retinal ganglion cells.
\newblock \emph{PLoS One}, 16\penalty0 (2):\penalty0 e0246952, 2021.

\bibitem[Rentzeperis et~al.(2014)Rentzeperis, Nikolaev, Kiper, and van Leeuwen]{rentzeperis2014distributed}
Ilias Rentzeperis, Andrey~R Nikolaev, Daniel~C Kiper, and Cees van Leeuwen.
\newblock Distributed processing of color and form in the visual cortex.
\newblock \emph{Frontiers in psychology}, 5:\penalty0 932, 2014.

\bibitem[Rezeanu et~al.(2021)Rezeanu, Kuchenbecker, Barborek, Mazzaferri, Neitz, and Neitz]{rezeanu2021explaining}
Dragos~L Rezeanu, James Kuchenbecker, Rachel Barborek, Marcus Mazzaferri, Maureen Neitz, and Jay Neitz.
\newblock Explaining the absence of functional tetrachromacy in females with four cone types.
\newblock \emph{Investigative Ophthalmology \& Visual Science}, 62\penalty0 (8):\penalty0 527--527, 2021.

\bibitem[Rezende \& Mohamed(2015)Rezende and Mohamed]{rezende2015variational}
Danilo Rezende and Shakir Mohamed.
\newblock Variational inference with normalizing flows.
\newblock In \emph{International conference on machine learning}, pp.\  1530--1538. PMLR, 2015.

\bibitem[Riesen et~al.(1964)Riesen, Ramsey, and Wilson]{riesen1964development}
Austin~H Riesen, Robert~L Ramsey, and Paul~D Wilson.
\newblock Development of visual acuity in rhesus monkeys deprived of patterned light during early infancy.
\newblock \emph{Psychonomic Science}, 1\penalty0 (1-12):\penalty0 33--34, 1964.

\bibitem[Rigotti et~al.(2013)Rigotti, Barak, Warden, Wang, Daw, Miller, and Fusi]{rigotti2013importance}
Mattia Rigotti, Omri Barak, Melissa~R Warden, Xiao-Jing Wang, Nathaniel~D Daw, Earl~K Miller, and Stefano Fusi.
\newblock The importance of mixed selectivity in complex cognitive tasks.
\newblock \emph{Nature}, 497\penalty0 (7451):\penalty0 585--590, 2013.

\bibitem[Rodieck(1965)]{rodieck1965quantitative}
Robert~W Rodieck.
\newblock Quantitative analysis of cat retinal ganglion cell response to visual stimuli.
\newblock \emph{Vision research}, 5\penalty0 (12):\penalty0 583--601, 1965.

\bibitem[Rodieck(1998)]{rodieck1998first}
Robert~W Rodieck.
\newblock The first steps in seeing.
\newblock 1998.

\bibitem[Rolfs(2009)]{rolfs2009microsaccades}
Martin Rolfs.
\newblock Microsaccades: small steps on a long way.
\newblock \emph{Vision research}, 49\penalty0 (20):\penalty0 2415--2441, 2009.

\bibitem[Ronneberger et~al.(2015)Ronneberger, Fischer, and Brox]{ronneberger2015u}
Olaf Ronneberger, Philipp Fischer, and Thomas Brox.
\newblock U-net: Convolutional networks for biomedical image segmentation.
\newblock In \emph{Medical Image Computing and Computer-Assisted Intervention--MICCAI 2015: 18th International Conference, Munich, Germany, October 5-9, 2015, Proceedings, Part III 18}, pp.\  234--241. Springer, 2015.

\bibitem[Roorda et~al.(2002)Roorda, Romero-Borja, Donnelly~III, Queener, Hebert, and Campbell]{roorda2002adaptive}
Austin Roorda, Fernando Romero-Borja, William~J Donnelly~III, Hope Queener, Thomas~J Hebert, and Melanie~CW Campbell.
\newblock Adaptive optics scanning laser ophthalmoscopy.
\newblock \emph{Optics express}, 10\penalty0 (9):\penalty0 405--412, 2002.

\bibitem[Rucci \& Victor(2015)Rucci and Victor]{rucci2015unsteady}
Michele Rucci and Jonathan~D Victor.
\newblock The unsteady eye: an information-processing stage, not a bug.
\newblock \emph{Trends in neurosciences}, 38\penalty0 (4):\penalty0 195--206, 2015.

\bibitem[Rucci et~al.(2018)Rucci, Ahissar, and Burr]{rucci2018temporal}
Michele Rucci, Ehud Ahissar, and David Burr.
\newblock Temporal coding of visual space.
\newblock \emph{Trends in cognitive sciences}, 22\penalty0 (10):\penalty0 883--895, 2018.

\bibitem[Sabesan et~al.(2015)Sabesan, Hofer, and Roorda]{sabesan2015characterizing}
Ramkumar Sabesan, Heidi Hofer, and Austin Roorda.
\newblock Characterizing the human cone photoreceptor mosaic via dynamic photopigment densitometry.
\newblock \emph{PloS one}, 10\penalty0 (12):\penalty0 e0144891, 2015.

\bibitem[Sabesan et~al.(2016)Sabesan, Schmidt, Tuten, and Roorda]{sabesan2016elementary}
Ramkumar Sabesan, Brian~P Schmidt, William~S Tuten, and Austin Roorda.
\newblock The elementary representation of spatial and color vision in the human retina.
\newblock \emph{Science advances}, 2\penalty0 (9):\penalty0 e1600797, 2016.

\bibitem[Salisbury \& Palmer(2016)Salisbury and Palmer]{salisbury2016optimal}
Jared~M Salisbury and Stephanie~E Palmer.
\newblock Optimal prediction in the retina and natural motion statistics.
\newblock \emph{Journal of Statistical Physics}, 162\penalty0 (5):\penalty0 1309--1323, 2016.

\bibitem[Schneidman et~al.(2006)Schneidman, Berry, Segev, and Bialek]{schneidman2006weak}
Elad Schneidman, Michael~J Berry, Ronen Segev, and William Bialek.
\newblock Weak pairwise correlations imply strongly correlated network states in a neural population.
\newblock \emph{Nature}, 440\penalty0 (7087):\penalty0 1007--1012, 2006.

\bibitem[Schottdorf \& Lee(2021)Schottdorf and Lee]{schottdorf2021quantitative}
Manuel Schottdorf and Barry~B Lee.
\newblock A quantitative description of macaque ganglion cell responses to natural scenes: the interplay of time and space.
\newblock \emph{The Journal of physiology}, 599\penalty0 (12):\penalty0 3169--3193, 2021.

\bibitem[Sibille et~al.(2022)Sibille, Gehr, Benichov, Balasubramanian, Teh, Lupashina, Vallentin, and Kremkow]{sibille2022high}
J{\'e}r{\'e}mie Sibille, Carolin Gehr, Jonathan~I Benichov, Hymavathy Balasubramanian, Kai~Lun Teh, Tatiana Lupashina, Daniela Vallentin, and Jens Kremkow.
\newblock High-density electrode recordings reveal strong and specific connections between retinal ganglion cells and midbrain neurons.
\newblock \emph{Nature communications}, 13\penalty0 (1):\penalty0 5218, 2022.

\bibitem[Singer et~al.(2018)Singer, Teramoto, Willmore, Schnupp, King, and Harper]{singer2018sensory}
Yosef Singer, Yayoi Teramoto, Ben~DB Willmore, Jan~WH Schnupp, Andrew~J King, and Nicol~S Harper.
\newblock Sensory cortex is optimized for prediction of future input.
\newblock \emph{elife}, 7:\penalty0 e31557, 2018.

\bibitem[Singer et~al.(2023{\natexlab{a}})Singer, Taylor, Willmore, King, and Harper]{10.7554/eLife.52599}
Yosef Singer, Luke Taylor, Ben~DB Willmore, Andrew~J King, and Nicol~S Harper.
\newblock Hierarchical temporal prediction captures motion processing along the visual pathway.
\newblock \emph{eLife}, 12:\penalty0 e52599, 2023{\natexlab{a}}.

\bibitem[Singer et~al.(2023{\natexlab{b}})Singer, Taylor, Willmore, King, and Harper]{singer2023hierarchical}
Yosef Singer, Luke Taylor, Ben~DB Willmore, Andrew~J King, and Nicol~S Harper.
\newblock Hierarchical temporal prediction captures motion processing along the visual pathway.
\newblock \emph{Elife}, 12:\penalty0 e52599, 2023{\natexlab{b}}.

\bibitem[Singh et~al.(2003)Singh, Freeman, and Brainard]{singh2003exploiting}
Barun Singh, William~T Freeman, and D~Brainard.
\newblock Exploiting spatial and spectral image regularities for color constancy.
\newblock In \emph{Workshop on Statistical and Computational Theories of Vision. Nice, France}. Citeseer, 2003.

\bibitem[Soto et~al.(2020)Soto, Hsiang, Rajagopal, Piggott, Harocopos, Couch, Custer, Morgan, and Kerschensteiner]{soto2020efficient}
Florentina Soto, Jen-Chun Hsiang, Rithwick Rajagopal, Kisha Piggott, George~J Harocopos, Steven~M Couch, Philip Custer, Josh~L Morgan, and Daniel Kerschensteiner.
\newblock Efficient coding by midget and parasol ganglion cells in the human retina.
\newblock \emph{Neuron}, 107\penalty0 (4):\penalty0 656--666, 2020.

\bibitem[Srinivasan et~al.(1982)Srinivasan, Laughlin, and Dubs]{srinivasan1982predictive}
Mandyam~Veerambudi Srinivasan, Simon~Barry Laughlin, and Andreas Dubs.
\newblock Predictive coding: a fresh view of inhibition in the retina.
\newblock \emph{Proceedings of the Royal Society of London. Series B. Biological Sciences}, 216\penalty0 (1205):\penalty0 427--459, 1982.

\bibitem[Stiles \& Burch(1955)Stiles and Burch]{stiles1955interim}
WS~Stiles and JM~Burch.
\newblock Interim report to the commission internationale de l'eclairage, zurich, 1955, on the national physical laboratory's investigation of colour-matching (1955).
\newblock \emph{Optica Acta: International Journal of Optics}, 2\penalty0 (4):\penalty0 168--181, 1955.

\bibitem[Stimper et~al.(2023)Stimper, Liu, Campbell, Berenz, Ryll, Schölkopf, and Hernández-Lobato]{Stimper2023}
Vincent Stimper, David Liu, Andrew Campbell, Vincent Berenz, Lukas Ryll, Bernhard Schölkopf, and José~Miguel Hernández-Lobato.
\newblock normflows: A pytorch package for normalizing flows.
\newblock \emph{Journal of Open Source Software}, 8\penalty0 (86):\penalty0 5361, 2023.
\newblock \doi{10.21105/joss.05361}.
\newblock URL \url{https://doi.org/10.21105/joss.05361}.

\bibitem[Stockman \& Sharpe(2000)Stockman and Sharpe]{stockman2000spectral}
Andrew Stockman and Lindsay~T Sharpe.
\newblock The spectral sensitivities of the middle-and long-wavelength-sensitive cones derived from measurements in observers of known genotype.
\newblock \emph{Vision research}, 40\penalty0 (13):\penalty0 1711--1737, 2000.

\bibitem[Stockman et~al.(1999)Stockman, Sharpe, and Fach]{stockman_spectral_1999}
Andrew Stockman, Lindsay~T. Sharpe, and Clemens Fach.
\newblock The spectral sensitivity of the human short-wavelength sensitive cones derived from thresholds and color matches.
\newblock \emph{Vision Research}, 39\penalty0 (17):\penalty0 2901--2927, 1999.
\newblock ISSN 0042-6989.

\bibitem[Stockman et~al.(2010)Stockman, Brainard, et~al.]{stockman2010color}
Andrew Stockman, David~H Brainard, et~al.
\newblock Color vision mechanisms.
\newblock \emph{OSA handbook of optics}, 3:\penalty0 11--1, 2010.

\bibitem[Tootell et~al.(1982)Tootell, Silverman, Switkes, and De~Valois]{tootell1982deoxyglucose}
Roger~BH Tootell, Martin~S Silverman, Eugene Switkes, and Russell~L De~Valois.
\newblock Deoxyglucose analysis of retinotopic organization in primate striate cortex.
\newblock \emph{Science}, 218\penalty0 (4575):\penalty0 902--904, 1982.

\bibitem[Torralba \& Oliva(2003)Torralba and Oliva]{torralba2003statistics}
Antonio Torralba and Aude Oliva.
\newblock Statistics of natural image categories.
\newblock \emph{Network: computation in neural systems}, 14\penalty0 (3):\penalty0 391, 2003.

\bibitem[Vaswani et~al.(2017)Vaswani, Shazeer, Parmar, Uszkoreit, Jones, Gomez, Kaiser, and Polosukhin]{vaswani2017attention}
Ashish Vaswani, Noam Shazeer, Niki Parmar, Jakob Uszkoreit, Llion Jones, Aidan~N Gomez, {\L}ukasz Kaiser, and Illia Polosukhin.
\newblock Attention is all you need.
\newblock \emph{Advances in neural information processing systems}, 30, 2017.

\bibitem[Verweij et~al.(2003)Verweij, Hornstein, and Schnapf]{verweij2003surround}
Jan Verweij, Eric~P Hornstein, and Julie~L Schnapf.
\newblock Surround antagonism in macaque cone photoreceptors.
\newblock \emph{Journal of Neuroscience}, 23\penalty0 (32):\penalty0 10249--10257, 2003.

\bibitem[Von~Helmholtz(1867)]{von1867handbuch}
Hermann Von~Helmholtz.
\newblock \emph{Handbuch der physiologischen Optik}, volume~9.
\newblock Voss, 1867.

\bibitem[Von~Kries(1902)]{von1902theoretische}
J~Von~Kries.
\newblock Theoretische studien {\"u}ber die umstimmung des sehorgans.
\newblock \emph{Festschrift der Albrecht-Ludwigs-Universit{\"a}t}, 32:\penalty0 145--158, 1902.

\bibitem[Wachtler et~al.(2007)Wachtler, Doi, Lee, and Sejnowski]{wachtler2007cone}
Thomas Wachtler, Eizaburo Doi, Te-Won Lee, and Terrence~J Sejnowski.
\newblock Cone selectivity derived from the responses of the retinal cone mosaic to natural scenes.
\newblock \emph{Journal of vision}, 7\penalty0 (8):\penalty0 6--6, 2007.

\bibitem[WALD(1968)]{wald1968molecular}
GEORGE WALD.
\newblock The molecular basis of visual excitation.
\newblock \emph{Nature}, 219\penalty0 (5156):\penalty0 800--807, 1968.

\bibitem[Wang et~al.(2019)Wang, Bensaid, Tiruveedhula, Ma, Ravikumar, and Roorda]{wang2019human}
Yiyi Wang, Nicolas Bensaid, Pavan Tiruveedhula, Jianqiang Ma, Sowmya Ravikumar, and Austin Roorda.
\newblock Human foveal cone photoreceptor topography and its dependence on eye length.
\newblock \emph{Elife}, 8:\penalty0 e47148, 2019.

\bibitem[Werner et~al.(2020)Werner, Marsh-Armstrong, and Knoblauch]{werner2020adaptive}
John~S Werner, Brennan Marsh-Armstrong, and Kenneth Knoblauch.
\newblock Adaptive changes in color vision from long-term filter usage in anomalous but not normal trichromacy.
\newblock \emph{Current Biology}, 30\penalty0 (15):\penalty0 3011--3015, 2020.

\bibitem[Williams et~al.(1981)Williams, MacLeod, and Hayhoe]{williams1981foveal}
David~R Williams, Donald~IA MacLeod, and Mary~M Hayhoe.
\newblock Foveal tritanopia.
\newblock \emph{Vision Research}, 21\penalty0 (9):\penalty0 1341--1356, 1981.

\bibitem[Williams(1983)]{williams1983pyramidal}
Lance Williams.
\newblock Pyramidal parametrics.
\newblock In \emph{Proceedings of the 10th annual conference on Computer graphics and interactive techniques}, pp.\  1--11, 1983.

\bibitem[Wool et~al.(2018)Wool, Crook, Troy, Packer, Zaidi, and Dacey]{wool2018nonselective}
Lauren~E Wool, Joanna~D Crook, John~B Troy, Orin~S Packer, Qasim Zaidi, and Dennis~M Dacey.
\newblock Nonselective wiring accounts for red-green opponency in midget ganglion cells of the primate retina.
\newblock \emph{Journal of Neuroscience}, 38\penalty0 (6):\penalty0 1520--1540, 2018.

\bibitem[Wright(1929)]{wright1929re}
William~David Wright.
\newblock A re-determination of the trichromatic coefficients of the spectral colours.
\newblock \emph{Transactions of the Optical Society}, 30\penalty0 (4):\penalty0 141, 1929.

\bibitem[Wu et~al.(2022)Wu, Brackbill, Sher, Litke, Simoncelli, and Chichilnisky]{wu2022maximum}
Eric Wu, Nora Brackbill, Alexander Sher, Alan Litke, Eero Simoncelli, and EJ~Chichilnisky.
\newblock Maximum a posteriori natural scene reconstruction from retinal ganglion cells with deep denoiser priors.
\newblock \emph{Advances in Neural Information Processing Systems}, 35:\penalty0 27212--27224, 2022.

\bibitem[Wyszecki \& Stiles(1982)Wyszecki and Stiles]{wyszecki1982color}
G{\"u}nther Wyszecki and Walter~Stanley Stiles.
\newblock \emph{Color science: concepts and methods, quantitative data and formulae}.
\newblock John wiley \& sons, 1982.

\bibitem[Yong(2022)]{yong2022immense}
Ed~Yong.
\newblock \emph{An immense world: How animal senses reveal the hidden realms around us}.
\newblock Knopf Canada, 2022.

\bibitem[Young \& Smithson(2021)Young and Smithson]{young2021emulated}
Laura~K Young and Hannah~E Smithson.
\newblock Emulated retinal image capture (erica) to test, train and validate processing of retinal images.
\newblock \emph{Scientific Reports}, 11\penalty0 (1):\penalty0 11225, 2021.

\bibitem[Young(1801)]{young1801ii}
Thomas Young.
\newblock Ii. the bakerian lecture. on the mechanism of the eye.
\newblock \emph{Philosophical Transactions of the Royal Society of London}, \penalty0 (91):\penalty0 23--88, 1801.

\bibitem[Young(1802)]{young1802ii}
Thomas Young.
\newblock Ii. the bakerian lecture. on the theory of light and colours.
\newblock \emph{Philosophical transactions of the Royal Society of London}, \penalty0 (92):\penalty0 12--48, 1802.

\bibitem[Zeki(1978)]{zeki1978functional}
Semir~M Zeki.
\newblock Functional specialisation in the visual cortex of the rhesus monkey.
\newblock \emph{Nature}, 274\penalty0 (5670):\penalty0 423--428, 1978.

\bibitem[Zhang et~al.(2022)Zhang, Cottaris, and Brainard]{zhang2022image}
Ling-Qi Zhang, Nicolas~P Cottaris, and David~H Brainard.
\newblock An image reconstruction framework for characterizing initial visual encoding.
\newblock \emph{Elife}, 11:\penalty0 e71132, 2022.

\bibitem[Zhang et~al.(2017)Zhang, Deng, Du, Zhu, Li, Xu, Sun, Gerstner, Baehr, Boye, et~al.]{zhang2017gene}
Yuxin Zhang, Wen-Tao Deng, Wei Du, Ping Zhu, Jie Li, Fan Xu, Jingfen Sun, Cecilia~D Gerstner, Wolfgang Baehr, Sanford~L Boye, et~al.
\newblock Gene-based therapy in a mouse model of blue cone monochromacy.
\newblock \emph{Scientific reports}, 7\penalty0 (1):\penalty0 6690, 2017.

\bibitem[Zhao et~al.(2023)Zhao, Ahissar, Victor, and Rucci]{zhao2023inferring}
Zhetuo Zhao, Ehud Ahissar, Jonathan~D Victor, and Michele Rucci.
\newblock Inferring visual space from ultra-fine extra-retinal knowledge of gaze position.
\newblock \emph{Nature communications}, 14\penalty0 (1):\penalty0 269, 2023.

\end{thebibliography}
